 \documentclass[final,5p,times,twocolumn]{elsarticle}

\usepackage{amssymb}
\usepackage{amsmath}
\usepackage[ruled]{algorithm2e}
\usepackage{comment}
\usepackage{multirow}
\usepackage{tabularx}
\usepackage{booktabs}
\usepackage{xcolor}
\usepackage{caption}
\usepackage{url}

\SetKwComment{Comment}{/* }{ */}

\usepackage{multirow}
\usepackage{lscape}
\usepackage{subfigure}

\journal{Neurocomputing}

\begin{document}

\begin{frontmatter}



\title{Compact Bayesian Neural Networks via pruned  MCMC sampling}


\author[inst4,inst3,inst1]{*Ratneel Deo}

\affiliation[inst4]{organization={ITTC ARC Centre for Data Analytics for Resources and Environment},
            addressline={ University of Sydney}, 
            city={Sydney},
            country={Australia}}

\author[inst2]{Scott Sisson}
            \affiliation[inst2]{organization={UNSW Data Science Hub},
            addressline={\& School of Mathematics and Statistics, University of New South Wales}, 
            city={Sydney},
            country={Australia}}

\author[inst3]{Jody M. Webster}
\affiliation[inst3]{organization={Geocoastal Research Group},
            addressline={ School of Geosciences, University of Sydney}, 
            city={Sydney},
            country={Australia}}
\author[inst4,inst1]{Rohitash Chandra}

\affiliation[inst1]{organization={Transitional Artificial Intelligence Research Group},
            addressline={School of Mathematics and Statistics, University of New South Wales}, 
            city={Sydney},
            country={Asutralia}}
            
\begin{abstract}
 Bayesian Neural Networks (BNNs) offer robust uncertainty quantification in model predictions, but training them presents a significant computational challenge. This is mainly due to the problem of sampling multimodal posterior distributions using  Markov Chain Monte Carlo (MCMC) sampling and variational inference algorithms. Moreover, the number of model parameters scales exponentially with additional hidden layers, neurons, and features in the dataset. Typically, a significant portion of these densely connected parameters are redundant and pruning a neural network not only improves portability but also has the potential for better generalisation capabilities.  In this study,  we address some of the challenges by leveraging MCMC sampling with network pruning to obtain compact probabilistic models having removed redundant parameters.  We sample the posterior distribution of model parameters  (weights and biases)  and prune weights with low importance, resulting in a compact model. We ensure that the compact BNN  retains its ability to estimate uncertainty via the posterior distribution while retaining the model training and generalisation performance accuracy by adapting post-pruning resampling. We evaluate the effectiveness of our MCMC pruning strategy on selected benchmark datasets for regression and classification problems through empirical result analysis.  We also consider two coral reef drill-core lithology classification datasets to test the robustness of the pruning model in complex real-world datasets. We further investigate if refining compact BNN can retain any loss of performance. Our results demonstrate the feasibility of training and pruning BNNs using MCMC whilst retaining generalisation performance with over 75\% reduction in network size. This paves the way for developing compact BNN models that provide uncertainty estimates for real-world applications.
 
\end{abstract}




\end{frontmatter}


\section{Introduction}




Bayesian neural networks (BNNs) and Bayesian deep learning (BDL) allow for the estimation of uncertainties in predictions, providing a more comprehensive and nuanced understanding of the model's output \cite{abdar_review_2021,chandra_bayesian_2024,chandra_langevin-gradient_2019}. This capability is crucial for real-world applications where uncertainty quantification is paramount, and the probabilistic nature of BNNs enhances interpretability by providing insights into the network's decision-making process.   BNNs utilise Bayesian inference, a probabilistic framework that updates the probability of a hypothesis as new evidence becomes available. It has been widely used in statistical models for parameter estimation, hypothesis testing, and decision-making  \cite{gelman_bayesian_1995, van_de_schoot_bayesian_2021}.  Markov Chain Monte Carlo (MCMC) sampling \cite{robert_metropolishastings_2004,roberts_exponential_1996} and variational inference \cite{jordan_introduction_1999} methods have been used to implement Bayesian inference. These methods have facilitated sampling (training) of robust BNN models \cite{neal_bayesian_2012, chandra_bayesian_2024,gal_bayesian_2016}, and have been capable of adapting to data that are either limited or noisy \cite{louizos_multiplicative_2017, nguyen_sequential_2024}.  


Although BNNs have shown promise in addressing issues such as uncertainty quantification and data scarcity, their adoption in mainstream off-the-shelf models remains limited. This is mainly because BNNs have challenges with convergence and scalability, especially with larger datasets \cite{papamarkou_challenges_2022, pall_bayesreef_2020, chandra_langevin-gradient_2019, trippe_overpruning_2018}.


Pruning neural networks and deep learning models  \cite{liang_pruning_2021, he_structured_2024} has been a crucial strategy for removing inessential model parameters without any significant effect on the performance \cite{sietsma_neural_1988, rawat_deep_2017, he_structured_2024, blalock_what_2020}. Han et al. \cite{han_learning_2015} demonstrated that a significant portion of weights can be set to zero without a significant loss in performance through additional L1/L2 loss function and gradual network pruning.  Pruning methods exhibit significant variability in their approaches, such as structural matrices \cite{sindhwani_structured_2015}, model quantization \cite{wu_quantized_2016}, model binarization \cite{courbariaux_binaryconnect_2015}, and parameter sharing \cite{han_learning_2015, wang_packing_2019}.  Unstructured pruning, also referred to as, parameter-based pruning, targets individual network parameters, resulting in sparse neural networks with a reduced parameter count\cite{sietsma_neural_1988, blalock_what_2020}. Conversely, structured pruning considers groups of parameters, such as entire neurons, filters, or channels, thus leveraging hardware and software enhancement for dense computation and achieving faster training performance\cite{li_pruning_2017, he_channel_2017}. 


Scoring-based pruning methods use the absolute values of parameters, trained importance coefficients, or contributions to network activations or gradients. Han et al. \cite{han_learning_2015} applied local scoring and pruned a fraction of the lowest-scoring parameters within each structural subcomponent of NNs.   There has been work on employed early pruning using \textit{lottery ticket hypothesis} on deep neural network architectures, achieving faster training and equivalent predictive performance when compared to unpruned dense networks \cite{lee_snip_2018, frankle_stabilizing_2020}.  Liu et al. \cite{liu_rethinking_2019} used scheduling-based pruning strategies to prune all desired weights in a single step, while  \cite{han_learning_2015}  iteratively pruned a fixed fraction over several steps. Recent advancements in network pruning literature have introduced dynamic pruning methods that remove parameters during training. \cite{gale_state_2019}  used magnitude pruning to adjust the pruning rate throughout the training process. Evolutionary algorithms have also gained traction for neural network pruning, allowing for iterative modifications to the network structure based on an initial sparse topology \cite{onan_hybrid_2017, fernandes_jr_pruning_2021}. However, these structured pruning methods typically focus on optimizing network performance without considering uncertainty in the parameters, which is the focus of our study.

 Although not directly related, there has been some work in the area of \textit{knowledge distillation} using Bayesian neural networks and deep learning.  Knowledge Distillation \cite{gou_knowledge_2021} provides the means of transferring the knowledge encapsulated in a large and complex deep learning model (the teacher) to a smaller, more efficient model (the student) \cite{schmidhuber_learning_1992, gou_knowledge_2021, yim_gift_2017}. Knowledge Distillation can be seen as an approach for reducing network (model) size and complexity.   Schmidhuber \cite{schmidhuber_learning_1992} was the first to demonstrate an instance of the knowledge distillation process through the compression of Recurrent Neural Networks (RNNs). This was later generalised by   Hinton et al. \cite{hinton_distilling_2015}, showcasing its effectiveness in image classification tasks. Knowledge distillation has since been applied to various domains, including object detection \cite{chen_learning_2017, li_when_2023}, acoustic modelling \cite{takashima_investigation_2018, asami_domain_2017}, and natural language processing \cite{fu_lrc-bert_2021, jiao_tinybert_2019}, demonstrating its versatility and utility in enhancing the performance and efficiency of smaller models across a range of applications.  Therefore, knowledge distillation and neural network pruning reduce model complexity while maintaining performance, which facilitates efficient deployment in resource-constrained environments.

 Regularisation techniques have been designed to improve generalisation performance, but they can also reduce the complexity of neural networks by creating simpler models that perform well on both training and test data, thereby improving robustness. Regularisation techniques such as dropout \cite{n_dropout_2014} have become integral to training deep neural networks, preventing overfitting and improving model generalisation, while reducing model complexity. Dropout regularisation stochastically deactivates a fraction of neurons during training, effectively introducing noise and encouraging the network (model) to learn more robust features. This technique has been widely adopted in various neural network architectures, including deep learning models such as Convolutional Neural Networks (CNNs) \cite{wu_towards_2015, park_analysis_2017}, RNNs \cite{gal_theoretically_2016, pham_dropout_2014}, and Large Language Models (LLMs)  \cite{lee_mixout_2019, yang_ad-drop_2022}. However, we note that the conventional dropout strategy is implemented only once, either before or after training to get a sparse network.  This may lead to discrepancies between training and inference phases, as no post-pruning model verification is done \cite{li_survey_2023}. Gal et al. \cite{gal_dropout_2016} framed dropout as a Bayesian approximation and used a variational inference with a Gaussian process to address this problem. The authors argued that deep learning models can converge to a finite Gaussian process as dropout is applied.   Hron et al. \cite{hron_variational_2018} reviewed some of the pitfalls of using variational dropouts, having improper priors and singular distribution, and proposed alterations that use quasi-KL divergence to work with dropouts in optimisation-based approximate inference algorithms. Graves \cite{graves_practical_2011} proposed a stochastic variational method for pruning RNNs using a novel signal-to-noise pruning heuristic. The heuristic removed network weights with high probability density at zero.  The method provided less than 3\% post-pruning performance loss on automatic speech recognition tasks.

 BNNs have received less attention in pruning research, and some key studies are discussed as follows. Sum et al. \cite{sum_adaptive_1999} used extended Kalman Filters to create an adaptive Bayesian pruning to reduce the complexity of the network. Sharma et al.   \cite{sharma_bayesian_2021} created BPrune, an open-source software package for pruning  CNNs with re-parameterization of variational posterior using Bayes-by-backprop \cite{blundell_weight_2015}.  There have been attempts in pruning in BNNs trained by variational inference.  Tripe and Turner \cite{trippe_overpruning_2018} explored the counterintuitive phenomenon of variational overpruning by reviewing the impact of selecting variational families—such as weight noise and mean-field, and reported that expressive variational families perform worse than simpler ones. Beckers et al. \cite {beckers_principled_2023}  pruned BNNs based on Bayesian model reduction in a greedy and iterative fashion using variance backpropagation and Bayes-by-backprop.  

Although variational inference-based approaches in the Bayesian framework have explored pruning \cite{trippe_overpruning_2018, beckers_principled_2023}, MCMC-based pruning remains underexplored.  The limited work in this area motivates the need for further investigation into pruning methods tailored to the Bayesian setting, especially in the MCMC sampling context. Developing effective pruning strategies for BNNs in the MCMC framework could unlock their full potential, enhancing model efficiency and scalability while preserving uncertainty estimation capabilities crucial for real-world applications. This is mainly due to the problem of sampling multimodal posterior distributions using   MCMC  sampling and variational inference algorithms. Moreover, the number of model parameters scales exponentially with additional hidden layers and features in the dataset. Typically, a significant portion of these densely connected parameters are redundant and pruning a neural network not only improves portability but also has the potential for better generalisation capabilities.

Langevin MCMC sampling combines  Langevin dynamics with Bayesian inference to incorporate gradient information in the proposal distribution for effectively sampling BNN posteriors \cite{chandra_langevin-gradient_2019}.  Gu et al. \cite{gu_neural_2020} addressed the issue of high autocorrelation using neural networks with Langevin MCMC sampling using a novel sampler neural networks Langevin Monte Carlo (NNLMC) using a customised loss function not to break the detailed balance condition.  Chandra et al. \cite {chandra_langevin-gradient_2019} used parallel tempering MCMC to improve the efficiency of sampling BNN parameters for classification and regression problems.  Parayil et al. \cite{parayil_decentralized_2020}  used a decentralised Langevin MCMC for image classification with improved accuracy with enhanced speed of convergence over conventional MCMC sampling. Look et al. \cite{look_differential_2020} used differential BNNs using an adaptation of   Langevin MCMC for time series and regression problems. Gurbuzbalaban et al. \cite{gurbuzbalaban_decentralized_2021} applied a decentralised  Langevin MCMC for Bayesian linear and logistic regression tasks in a decentralised setting. Garriga-Alonso et al.  \cite{garriga-alonso_exact_2020} used a gradient-guided Monte Carlo sampler incorporating gradient-based proposal distribution that can be used with stochastic gradients, yielding nonzero acceptance probabilities computed across multiple steps for Bayesian deep learning. 

In this study,  we address some of the challenges faced by BNNs by leveraging MCMC sampling with network pruning for obtaining compact probabilistic models. Hence, we present a novel approach for BNN pruning using  MCMC sampling for compact model and uncertainty quantification.   We sample the posterior distribution of model parameters  (weights and biases)  and prune weights with low importance during the sampling process, resulting in a compact model. We ensure that the compact BNN  retains its ability to estimate uncertainty via the posterior distribution while maintaining the model training and generalisation performance accuracy. We evaluate the effectiveness of our  MCMC pruning strategy on selected benchmark datasets for regression and classification problems. We track samples' signal-to-noise ratio \cite{graves_practical_2011} in the posterior distribution of the weight/bias post-sampling to potentially eliminate them based on a pruning threshold.  This is done post-training to ensure adequate inference.  We look at selected regression and classification problems from the literature \cite {chandra_bayesian_2017} to measure pre- and post-pruning performance.  The problems are chosen to ensure a fair representation of synthetic and real-world datasets. We further investigate if refining compact BNN can retain any loss of performance.  Finally, we apply the methodology to a real-world problem of detecting lithologies in reef drill core data from the Great Barrier Reef.

The next section \ref{sec:background} provides a brief overview of the work done in pruning literature.  We give details of the proposed methodology in Section \ref{sec:method} with an overview of the datsets, model and results in Section \ref{sec:results}.  Finally, we provide a discussion and summarise our findings in Sections \ref{sec:discussion}  and \ref{sec:conclusion}, respectively.

\section{Background and Related Work}
\label{sec:background}

\subsection{Neural network pruning methods}

 Pruning neural networks have been effective as a means to avoid overfitting, which requires a trade-off between the models' performance and complexity  \cite{lecun_optimal_1989}. Pruning methods involve removing parameters (weights/biases) that do not contribute to the decision-making process and are deemed unnecessary. A naive approach assessed the importance of the parameters directly based on their value \cite{sietsma_neural_1988}, where the smallest parameters were regarded as unimportant and removed. However, it was later suggested that these parameters could be necessary to achieve a high model performance \cite{hassibi_second_1992}. Alternative methods rely on the Hessian of the loss function \cite{hassibi_second_1992}, or the gradients during training \cite{strom_phoneme_1997}. A sparse (compact) neural network is obtained when parameters are removed from a model. However, such methods require additional computation resources to calculate the pruning criteria \cite{han_learning_2015}, and more recent methods prune the network based on the parameter values and contributions. However, as opposed to the earlier methods where parameters are removed after training, they may be removed dynamically during training, using simple criteria such as the weight magnitude \cite{zhu_prune_2017, mostafa_parameter_2019}, momentum magnitude \cite{dettmers_sparse_2019} and the signal-to-noise ratio of mini-batch gradients \cite{siems_dynamic_2021}.

 Pruning and model compression techniques have been critical in the efficiency of deep learning models, including resource-constrained environments.  Zhu et al. \cite{zhu_prune_2017} reported that while reducing the number of model parameters can effectively decrease the model size and computational load, overly aggressive pruning can degrade performance, thus necessitating a careful balance.  Tung and Mori \cite{frederick_deep_2018} introduced a compression technique that combined pruning with quantisation, the process of reducing the numerical precision of weights. The study reported high model compression rates while maintaining model accuracy, making it viable for real-time applications where efficiency is paramount.  Hoefler et al. \cite{hoefler_sparsity_2021} reviewed various sparsity techniques and reported that dynamic sparsity, wherein neurons were pruned or grown adaptively based on relevance, provided faster inference and reduced memory use during training and deployment. Yeom et al. \cite{yeom_pruning_2021} developed a pruning criterion based on model interpretability, which leveraged weight importance interpreted through the model’s decision-making ability. The method enabled the model to become smaller and more explainable, which is valuable in applications where model transparency is crucial. Zemouri et al. \cite{zemouri_new_2020} proposed a pruning algorithm that dynamically adjusts model size by pruning or growing neurons during training based on task complexity. This flexibility enables better generalisation and adaptability to diverse problem domains. Vadera and Ameen \cite{vadera_methods_2022} provided a broad overview of pruning strategies, discussing the balance between model simplification and accuracy preservation. The study examined the efficiency and model performance of unstructured pruning that removed individual weights, and structured pruning that eliminated larger components such as filters.   


 Pruning techniques have been extensively studied for improving the efficiency of  CNNs \cite{he_structured_2024, wang_packing_2019, cheng_survey_2024}.  He and Xiao \cite{he_structured_2024} surveyed structured pruning techniques for CNNs, emphasising the importance of balancing model accuracy and efficiency.  Molchanov et al. \cite{molchanov_pruning_2017} proposed a Taylor expansion-based criterion for pruning convolutional kernels in CNNs.  The method demonstrated superior performance in transfer learning tasks and large-scale image classification datasets, achieving significant computational reductions. Anwar et al. \cite{anwar_structured_2017} developed a structured pruning approach for CNNs, where entire convolutional layer filters, rather than individual weights were pruned to allow for substantial reductions in both model size and computational complexity, making it suitable for embedded systems with limited resources. Tang et al. \cite{tang_reborn_2020} developed a reborn filter technique for pruning in scenarios with limited data by leveraging a filter re-initialisation strategy to recover model performance after pruning. Wang et al. \cite{wang_convolutional_2021} proposed structural redundancy reduction to strategically identify and remove redundant components in CNNs to improve efficiency. In general, these studies underline the growing sophistication of pruning techniques with the aim of balancing computational efficiency with predictive performance.


 Evolutionary computation methods have gained traction for pruning deep learning models over the past decade \cite{mantzaris_genetic_2011, poyatos_evoprunedeeptl_2023, stanley_evolving_2002}. These methods leverage evolutionary algorithms to optimise the selection of pruned weights, filters,  and layers. Stanley et al. \cite{stanley_evolving_2002} proposed the NeuroEvolution of Augmenting Topologies (NEAT) to prune probabilistic neural networks with a genetic algorithm.  This concept was later extended with pruning during training, where the algorithm modifies the network structure repeatedly at the end of each training epoch, based on an initial sparse topology. Cantú-Paz \cite{cantu-paz_pruning_2003} used simple genetic algorithms for pruning a feedforward neural network trained with standard backpropagation on benchmark classification datasets. Yang and Chen \cite{yang_evolutionary_2012} introduced a constructive and pruning evolutionary algorithm that adapts the neural network structure during training.  The model performed well in time series prediction problems. Fernandes and Yen \cite{fernandes_jr_pruning_2021} used evolutionary strategies for pruning deep convolutional neural networks, showing that such methods can reduce model complexity while preserving accuracy, thereby providing a valuable tool for designing efficient neural network architectures.  Samala et al.\cite {samala_evolutionary_2018} pruned CNNs trained for breast cancer diagnosis using a genetic algorithm without retraining and reported no statistically significant loss in performance when compared to the original model.  Zhou et al. \cite{zhou_evolutionary_2022}  proposed an evolutionary algorithm-based method for pruning deep neural networks using multiobjective optimization for removing blocks while maintaining network performance with classification on ImageNet datasets.  Poyatoes et al. \cite{poyatos_evoprunedeeptl_2023} developed an evolutionary pruning model for pruning transfer learning-based deep learning model that replaced the (last) fully connected layer with a sparse layer optimised by a genetic algorithm.


 Recent advances in pruning techniques for neural networks have demonstrated the applicability of Bayesian methods for principled model compression and regularisation. Williams \cite{williams_bayesian_1995} introduced Bayesian regularisation and pruning using a Laplace prior, which encouraged sparsity by penalising irrelevant weights, providing a foundation for Bayesian pruning methodologies. Neklyudov et al. \cite{neklyudov_structured_2017} proposed a Bayesian model for introducing structured sparsity in neural networks by employing a truncated log-uniform prior and log-normal variational approximation.  The model injected noise into neuron outputs while keeping weights unregularized, removing low signal-to-noise elements to accelerate deep architectures. Van Baalen et al. \cite{van_baalen_bayesian_2020} unified pruning and quantisation within a Bayesian framework for image classification tasks.  The method involved selecting priors for Bayesian inference by performing group sparsity on the output channels of the weights of deep learning models. Beckers et al. \cite{beckers_principled_2023} proposed a variational free energy minimisation approach to prune Bayesian neural networks, balancing model complexity and predictive performance while retaining uncertainty quantification. Mathew and Rowe \cite{mathew_pruning_2023} utilised Bayesian inference to evaluate weight importance, producing sparse models with minimal performance degradation. Collectively, these studies underscore the growing integration of Bayesian principles in pruning strategies, emphasising uncertainty quantification, computational efficiency, and adaptability to various neural network architectures. These studies have illustrated a spectrum of pruning methodologies that address the pressing need for efficient, and scalable neural networks in critical applications.

\subsection{Bayesian Inference for Neural Networks}

Bayesian inference via MCMC sampling estimates the posterior distribution based on the prior and the likelihood that takes into account the observed data. The posterior distribution represents the updated probability distribution of the parameters given the observed data. According to Bayes' theorem, the posterior distribution is proportional to the product of the likelihood and the prior:

\begin{equation}
p(\theta|y,x) \propto p(y|x, \theta) \cdot p(\theta)
\label{eq:posterior}
\end{equation}

Calculating the exact posterior distribution is often intractable due to the complex structure of neural networks. Therefore, approximate inference methods such as  MCMC sampling \cite{chib_understanding_1995, robert_metropolishastings_2004} and variational inference \cite{jordan_introduction_1999} have been used to approximate the posterior distribution.

BNNs \cite{gelman_bayesian_1995} take a probabilistic approach to the uncertainty quantification in parameters of neural network models. In contrast to simple neural networks and deep learning models, where parameters (weights and biases) are point estimates, BNNs represent them as probability distributions. The BNN features the probabilistic representation of weights and biases sampled (trained) using MCMC sampling or variational inference. In this study, we focus on Langevin MCMC sampling of BNNs. Given a dataset $\mathbf{D} = \{(\mathbf{x}_i, y_i)\}_{i=1}^N$  with input features ($\mathbf{x}_i$) and the corresponding output ($y_i$) having $N$ number of data points, we define a simple neural network model (one hidden layer) with a set of weights and biases, also known as the parameters ($\theta$). The model produces predictions $\hat{y}_i = f(\mathbf{x}_i; \theta)$, where $f(.)$ represents the neural network model. In BNNs, the parameters are considered random variables represented by user-defined probability distributions.   We build two separate BNN models with input data $\mathbf{x}$ and corresponding labels $y$.  Firstly, we build a regression model;
\begin{equation}
    y=f(\mathbf{x}, \theta ) + e    \qquad \qquad e \sim \mathcal{N}(0, \tau ^{2}) 
    \label{eq:model_regression}
\end{equation}
We assume the regression model is affected by Gaussian noise, so we add the error term $e$ that follows a Gaussian distribution with mean $0$ and a standard deviation of $\tau ^{2}$. We then postulate a likelihood function representing the probability of observing the data given the model's parameters. In the case of regression, it's often assumed to follow a Gaussian distribution.  

\begin{equation}
    p(y|\mathbf{x}, \theta) = \mathcal{N}(f(\mathbf{x}|\theta), \tau^2).
    \label{eq:likelihood_regression}
\end{equation}
where $f(\mathbf{x}|\theta)$ is the output of the neural network given input $\mathbf{x}$ and parameters $\theta$, and $\tau^2$ is the variance of the Gaussian noise.

The classification model is much simpler. We do not have any Gaussian error in the model as we have discrete outputs. 
\begin{equation}
    y=f(\mathbf{x}, \theta ) 
    \label{eq:model_classification}
\end{equation}
However, the model output for classification is represented by \( \mathbf{y} \). This is a vector representing the probability distribution over classes as we are working with multi-class classification.  Therefore, the likelihood function, in this case, is defined using a Multinomial distribution:

\begin{equation}
    p(\mathbf{y} | \mathbf{x}, \theta) = \text{Multinomial}(\text{softmax}(f(\mathbf{x} | \theta)))
    \label{eq:likelihood_classification}
\end{equation}

We use a prior distribution $P(\theta)$  for the parameters $\theta$, encoding initial beliefs about their distribution. Upon observing data $\mathbf{D}$, we can update our beliefs about $\theta$ by computing the posterior distribution $P(\theta | \mathbf{D})$ using Bayes' theorem:

\begin{equation*}
P(\theta | \mathbf{D}) = \frac{P(\mathbf{D} | \theta) P(\theta)}{P(\mathbf{D})} \propto P(\mathbf{D} | \theta) P(\theta),
\end{equation*}
where $P(\mathbf{D} | \theta)$ is the likelihood function, and $P(\mathbf{D})$ is the marginal likelihood of the data.

We note that a prior distribution in principal needs to be chosen before examining the data for the given problem \cite{vladimirova_understanding_2019}, and priors for the case of neural networks is a challenging problem given multimodal posterior distributions \cite{izmailov_what_2021}. We can get expert opinions about trained neural networks from the literature to specify a prior distribution for the model parameters. Typically, in the case of BNNs, the prior for the weights and biases \cite{chandra_bayesian_2024, chandra_langevin-gradient_2019} is a Gaussian distribution with zero mean and a specified variance 
\begin{equation}
    p(\theta) = \prod_{i=1}^T \mathcal{N}(0, \sigma_i^2)
    \label{eq:prior}
\end{equation}
where $T$ is the total number of model parameters  and $\sigma_i^2$ is the variance for the parameter indexed by $i$.

\subsection{Langevin Bayesian Neural Networks}

Langevin MCMC sampling \cite{roberts_exponential_1996}   is a sophisticated approach that combines gradient information to develop an effective proposal distribution for BNNs \cite{chandra_langevin-gradient_2019}.   Langevin MCMC sampling utilises gradient information computed via backpropagation to enhance the proposal distribution and demonstrated to improve both convergence speed and accuracy \cite{chandra_langevin-gradient_2019, chandra_revisiting_2022}.  The Langevin proposal distribution involves blending deterministic and stochastic components through Langevin dynamics. The deterministic component is the gradient computed for a given set of parameters ($\theta$) obtained from the gradient of the log-likelihood (log-posterior) through backpropagation. The gradient guides the sampling process that provides an adaptive proposal distribution rather than a fixed one as in the case of the standard random-walk proposal distribution.  The second component involves a stochastic noise term, typically a Gaussian, added to the gradient update which prevents the sampling from converging prematurely to local optima. This can also ensure that the parameter space is explored more thoroughly with better mixing. By combining these two components, we get the Langevin proposal distribution:

\begin{equation}
\theta' = \theta + \frac{\epsilon}{2} \nabla_\theta \log P(\theta|\mathbf{d}) + \eta,
\label{eq:langevin_dynamics}
\end{equation}
where $\epsilon$ is the step size, $\nabla_\theta \log P(\theta|\mathbf{d})$ is the gradient of the log posterior, and $\eta$ represents Gaussian noise. The proposal is then subjected to an acceptance criterion defined by the Metropolis-Hastings ratio, and upon acceptance, it becomes part of the chain for the posterior distribution of weights and biases.

Langevin MCMC sampling methods can reduce the number of iterations (samples), making Bayesian inference feasible even for high-dimensional deep-learning models. Although BNNs have been around for more than two decades, their implementation has been slow due to the challenge of sampling including a large number of posterior distributions in complex models. More recently, Bayesian deep learning models such as Bayesian autoencoders \cite{chandra_revisiting_2022} and Bayesian CNNs \cite{chandra_bayesian_2021} have been trained with MCMC sampling that employed gradient information, including Langevin and Hamiltonian MCMC sampling methods. Chandra et al. \cite{chandra_bayesian_2021} utilized tempered MCMC sampling with adaptive Langevin-gradient proposals for Bayesian CNNs on multi-class classification problems. Nguyen et al. \cite{nguyen_sequential_2024}  proposed a sequential reversible-jump MCMC for dynamic BNNs. The model sampled the network topology and the parameters in parallel to look for uncertainty in model structures with comparable performance on benchmark classification and regression problems.  Variational inference has been successfully used for Bayesian deep learning models, including CNNs \cite{gal_bayesian_2016, chandra_bayesian_2021} and RNNs \cite{kapoor_cyclone_2023}.  Hamiltonian MCMC uses Hamiltonian dynamics with momentum variables to efficiently explore posterior distributions, while Langevin MCMC uses gradient-based updates with stochastic noise, making HMC more structured but computationally intensive \cite{chandra_revisiting_2022}.  Furthermore,  Langevin MCMC sampling has demonstrated that the quality of posterior sampling improves since the gradient-informed updates ensure that samples concentrate around regions of high posterior probability, resulting in better uncertainty estimates through improved accuracy and reduced credible interval \cite{chandra_bayesian_2021, chandra_langevin-gradient_2019, chandra_bayesian_2024}.    




\subsection{Model and Likelihood in BNNs}

In Langevin MCMC sampling of BNNs, the model and likelihood functions are essential in defining the probabilistic framework.  We need to define the prior and likelihood functions differently for regression and classification tasks as the models are different, as given in Equation \ref{eq:model_regression} and \ref{eq:model_classification}.  We derive our likelihoods and prior from work previously done by \cite{chandra_bayesian_2017,chandra_langevin-gradient_2019,chandra_bayesian_2024} as the model and data we use are similar to these works. In the case of regression, we deal with continuous values in the prediction and will use a Gaussian likelihood function. Therefore, use a  Gaussian prior distribution of the weights and biases $\theta$ and an inverse-Gamma distribution over the noise variance $\tau^2$. In the case of classification problems, we will deal with discrete values in the prediction and use multinomial likelihood. In this case, we do not have the $\tau^2$ as a parameter that will be sampled and do not need a prior for it. The prior assumes a Gaussian distribution over the weights and biases $\theta$.  

We define the log-prior for regression  by:
\begin{equation}
\log P(\theta, \tau^2) = -\frac{T}{2} \log \sigma^2 - \frac{1}{2 \sigma^2} \sum_{i=1}^{T} \theta_i^2 - (1 + \nu_1) \log \tau^2 - \frac{\nu_2}{\tau^2},
\label{eq:prior_regression}
\end{equation}
where $\sigma^2$ is the Gaussian prior variance on weights,and $\nu_1$ and $\nu_2$ are hyperparameters for the inverse-Gamma prior distribution used for   $\tau^2$.  We define the log-prior for classification by using a Gaussian prior given by Equation \ref{eq:prior}:
\begin{equation}
\log P(\theta) = -\frac{T}{2} \log \sigma^2 - \frac{1}{2 \sigma^2} \sum_{i=1}^{T} \theta_i^2 .
\label{eq:prior_classification}
\end{equation}

The likelihood functions also differ between regression and classification tasks, with a Gaussian likelihood for regression and a multinomial likelihood for classification. For an individual data point $(\mathbf{x}_i, y_i)$, the Gaussian likelihood is:
\begin{equation*}
    P(y_i | \mathbf{x}_i, \theta) = \frac{1}{\sqrt{2 \pi \tau^2}} \exp \left( -\frac{(y_i - f(\mathbf{x}_i; \theta))^2}{2 \tau^2} \right),
\end{equation*}
where $f(\mathbf{x}_i; \theta)$ is the model prediction for input $\mathbf{x}_i$, $\theta$ are the model parameters (weights and biases), and $\tau^2$ is the variance of the observation noise. Therefore, for the entire dataset $\mathbf{D}$ with $N$ data points, the Gaussian log-likelihood is:
\begin{equation}
    \log P(\mathbf{D} | \theta) = -\frac{N}{2} \log(2 \pi \tau^2) - \frac{1}{2 \tau^2} \sum_{i=1}^N (y_i - f(\mathbf{x}_i; \theta))^2.
    \label{eq:logliklehood_regression}
\end{equation}

For classification tasks, where there are $K$ possible outcomes for each observation, the likelihood assumes a multinomial distribution, where each class has a probability $P_k$ associated with it given by a Softmax function. For an individual data point with label $y_i$ in one-hot encoded form (i.e., $y_{i,j} = 1$ if the class is $j$, and $0$ otherwise), the likelihood is:
\begin{equation}
y_i \backsim Multinomial (P_1, ..., P_K), 
\nonumber
\end{equation}
\begin{equation}
y_i \in \zeta , \zeta = (1,2,...,K), 
\nonumber
\end{equation}
\begin{equation}
Z_i = (y_{i,1},...,y_{i,K}),
\nonumber
\end{equation}
\begin{equation}
P(Z_i | \mathbf{x}_i, \theta) = \prod_{j=1}^K \left( p_j(\mathbf{x}_i; \theta) \right)^{Z_{i,j}},
\end{equation}
where $p_j(\mathbf{x}_i; \theta)$ is the probability that the neural network assigns to class $j$ for input $\mathbf{x}_i$.  For the entire dataset $\mathbf{D}$ with $N$ data points, the multinomial log-likelihood is:
\begin{equation}
\log P(\mathbf{D} | \theta) = \sum_{i=1}^N \sum_{j=1}^K Z_{i,j} \log p_j(\mathbf{x}_i; \theta).
\label{eq:loglikelihood_classification}
\end{equation}

The posterior distribution combines the prior information and the log-likelihood of the observed data to update our beliefs about the model parameters in regression and classification tasks. For regression tasks, with a Gaussian likelihood (Equation \ref{eq:logliklehood_regression}) and the regression prior (Equation \ref{eq:prior_regression}), the posterior distribution of the parameters \( \theta \) and noise variance \( \tau^2 \) given the data \( \mathbf{D} \) is:

\begin{equation}
P(\theta, \tau^2 | \mathbf{D}) \propto P(\mathbf{D} | \theta, \tau^2) P(\theta, \tau^2).
\nonumber
\end{equation}
\begin{align}
\log P(\theta, \tau^2 | \mathbf{D}) = & -\frac{N}{2} \log(2 \pi \tau^2) - \frac{1}{2 \tau^2} \sum_{i=1}^N (y_i - f(\theta))^2 \\
\nonumber
& -\frac{T}{2} \log \sigma^2 - \frac{1}{2 \sigma^2} \sum_{i=1}^{T} \theta_i^2 - (1 + \nu_1) \log \tau^2 - \frac{\nu_2}{\tau^2}.
\end{align}

For classification tasks, with a multinomial likelihood (Equation \ref{eq:loglikelihood_classification}) and the Gaussian prior (Equation \ref{eq:prior_classification}), the posterior distribution of the parameters \( \theta \) given the data \( \mathbf{D} \) is:
\begin{equation}
P(\theta | \mathbf{D}) \propto  P(\mathbf{D} | \theta) \cdot P(\theta),
\nonumber
\end{equation}
\begin{align}
 \log P(\theta | \mathbf{D}) =  &  \sum_{i=1}^N \sum_{j=1}^K Z_{i,j} \log p_j(\mathbf{x}_i; \theta) \\
 \nonumber
 & - \frac{1}{2\sigma^2} \sum_{i=1}^{T} \theta_i^2 + \text{Constant}.
\end{align}

\section{Methodology}
\label{sec:method}

\begin{figure*}[htb!]
    \centering
    \includegraphics[width=0.9\linewidth, height=\linewidth]{ 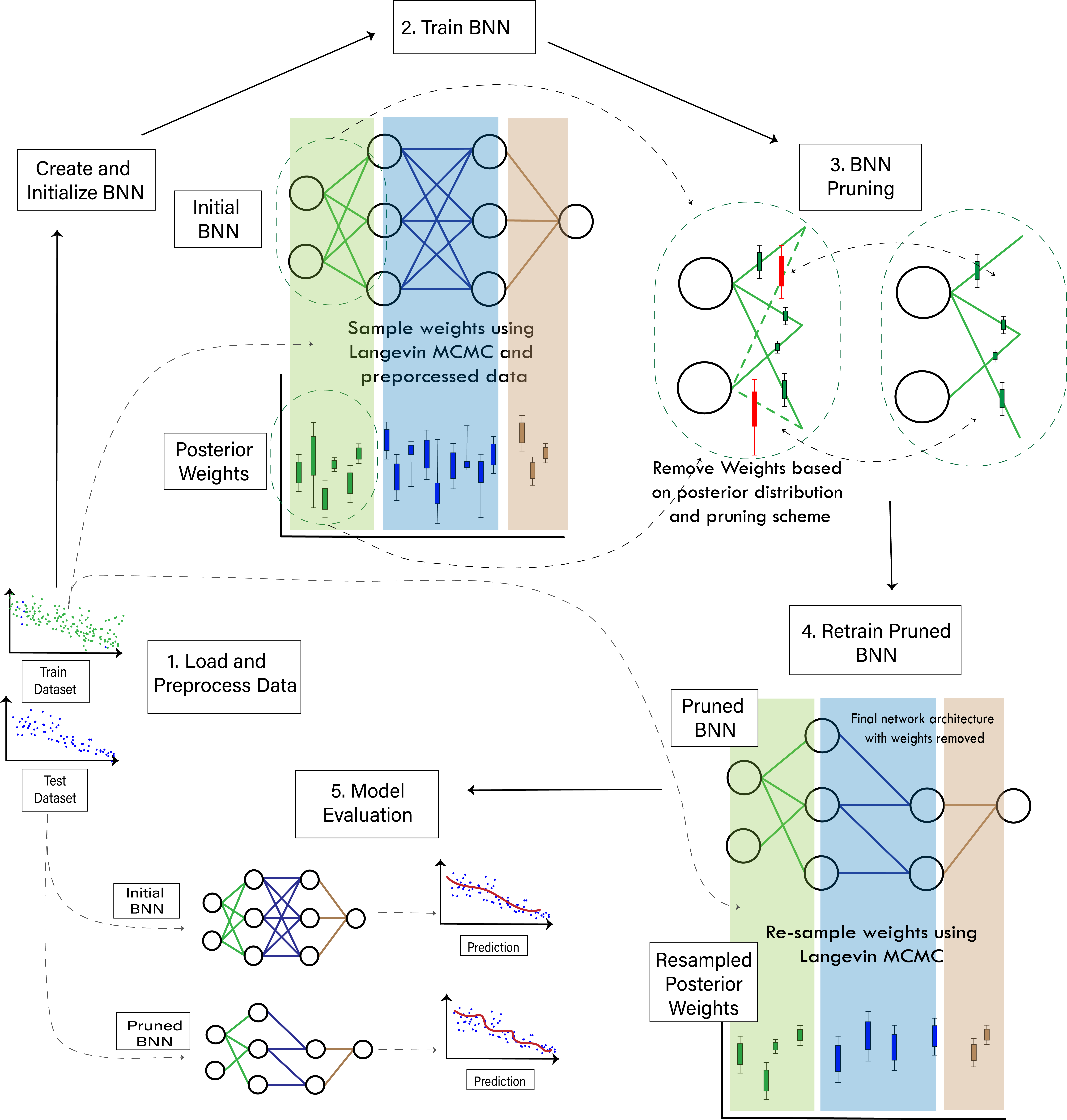}
    \caption{Framework for compact BNNs with network pruning post-sampling (training) where the weights/biases that do not contribute significantly to the posterior are removed. The compact BNN is later refined using the same training data to potentially regain the performance lost from pruning.  }
    \label{fig:framework}
\end{figure*}

\subsection{Pruning Algorithm} 

We propose a Bayesian pruning strategy that employs a signal-to-noise ratio inspired by Graves et al. \cite{graves_practical_2011} to get compact BNNs from   BNNs trained by Langevin MCMC sampling.  This approach has not been explored previously and offers a promising direction for enhancing model efficiency. In this way, we can identify and retain weights significantly contributing to the model's performance, effectively distinguishing valuable signals from background noise.  We focus on post-pruning, i.e., applying our strategy to a trained BNN  followed by a resampling phase via Langevin MCMC. 

Specifically, we consider two key pruning criteria: signal-to-noise \cite{graves_practical_2011} and signal-plus-noise \cite{nalisnick_priors_2018} ratio, and provide the details about their differences below. 

\subsubsection{Signal-to-Noise ratio}

Graves et al.  \cite{graves_practical_2011}  proposed the signal-to-noise (STN) ratio for pruning BNNs via variational inference with a Gaussian prior. Inspired by this, we apply it to pruning trained Langevin MCMC BNNs as given  by;

\begin{equation}
    \frac{|P\mu_i)|}{P\sigma_i} < \lambda,
    \label{eq:sigtonoise}
\end{equation}
where $P\mu_i$ and $P\sigma_i$ are the mean and standard deviation of the i-th weight of the model, respectively.  We select $\lambda$ as a constant that determines the threshold for pruning. We implement pruning by setting the weights to zero, for which this inequality holds. The weights with a much larger variance than the mean will be interpreted as 'noisy' and hence removed.

\subsubsection{Signal-plus-noise ratio}

We also use a slightly different criterion for pruning, called the signal-plus-noise  (SPN) ratio \cite{nalisnick_priors_2018}, defined by

\begin{equation}
    |P\mu_i| + P\sigma_i < \lambda,
    \label{eq:sigplusnoise}
\end{equation}
where $ P\mu_i$, $ P\sigma_i$, and $ lambda$ are same  as with SNR. Unlike SNR, SPN keeps the parameters whose mean and variances are both large and removes the ones whose means and variances are both small.

We need to assess the model performance after pruning for either criterion to determine the user-defined  $\lambda$. For each model, we prune the parameters with the lowest SPN and STN ratios.  We perform an inference to record the model's performance each time a weight has been pruned.  As a measure of robustness, we also randomly select weights and remove them from the BNN posterior, which we call random pruning in BNN.  This is to compare  STN and SPN  pruning and check if they have any benefits compared to the strategy where we randomly dropout selected weights/biases.

\begin{algorithm}[htb!]
    \caption{BNN  post-pruning strategy after training using Langevin MCMC sampling.}
    \label{alg:prunning}
    \KwData{Dataset}
    \KwResult{Pruned posterior distribution  of weights and biases}

    \textbf{Stage 1.0:} Preprocess data and initialise model \;
    Initialise $\theta_0$ \;
    Utilise train and test dataset \;
    
    \textbf{Stage 2.0:} Sample the posterior using Langevin MCMC\;
    \For{ $ i=1$ until $N_max$}{
        1. Propose a value $\theta'|\theta_i \sim q(x_i)$, where $q(.)$ is the proposal distribution\;
        2. Given $\theta$, execute the model $f(\theta; \textbf{d})$ to compute the predictions (output $y$) and the likelihood\;
        3. Calculate the acceptance probability \;
        $ \qquad \qquad \alpha = min ( 1, \frac{P(\theta')q(\theta_i|\theta)}{P(\theta_i)q(\theta"|\theta_i)})$ \;
        
        4. Generate a random value from a uniform distribution $\alpha \sim U(0, 1)$\;
        5. Accept or reject proposed value $\theta'$ \;
        
      \If{$\mu < \alpha$ }{
        $ \theta_i = \theta' $  \Comment*[r]{accept the sample} 
      }{\Else{
           $ \theta_i = \theta_{i-1} $  \Comment*[r]{reject the sample}
        }
      }
    }

    \textbf{Stage 3.0:} Pruning the weights/biases \;
    Sort the weights by the pruning criterion $\theta' = sorted(\theta)$ \;
    \For{ each $\theta_i$ in  $\theta'$}{
        \If{Pruning Ratio $ < \lambda $ }{
            Prune weights by setting $\theta_i \leftarrow 0$
        }
        \Else{
            Keep weight of $\theta_i $
        }
    }

    \textbf{Stage 4.0:} Resample existing weights/biases (retrain BNN)  \;
    Set $\theta_0$  based on  weights left over from pruning\;
    \For{ $ i=1$ until $Resample_duration$}{
        Using \textbf{Stage 2.0} resample the posterior
    }

    \textbf{Stage 5.0:} Predict with pruned BNN \;
    Evaluate the post-pruning performance of BNN \;
    
\end{algorithm}

Algorithm \ref{alg:prunning} outlines the method for pruning the BNN using Langevin MCMC sampling, followed by a pruning phase to reduce model complexity, which is also depicted in our framework (Figure \ref{fig:framework}). We obtain the compact BNN model by removing the low-impact weights/biases post-training.

We begin (Stage 1) by preprocessing the data and creating training and test sets as shown in the framework (Figure \ref{fig:framework}). In Stage 1, we initialise the  BNN  model based on the dataset and the likelihood function, which is dependent on the problem as given in the previous section. We draw and assign the initial model parameters ($\theta_0$) after defining model architecture such as the number of hidden neurons, and input and output neurons, which depend on the data. We prepare the model and data for posterior sampling, laying the groundwork for the MCMC process. We employ  Langevin MCMC sampling to iteratively construct the posterior distribution of the model parameters. In each iteration of the MCMC chain, we propose a new candidate set of parameters ($\theta_p$) and evaluate it using the likelihood function given in Equation \ref{eq:logliklehood_regression} or \ref{eq:loglikelihood_classification}. We either accept or reject the proposal based on an acceptance probability, $\alpha$, which is computed via the Metropolis-Hastings criterion as shown in Stage 2, Steps 4 and 5 of Algorithm 1. We need to ensure that the sampling process balances exploration and convergence towards high-probability regions of the posterior distribution. Therefore, the Metropolis-Hastings algorithm requires careful tuning of the proposal distribution to explore the parameter space while maintaining detailed balance efficiently, ensuring the Markov chain converges to the target posterior distribution \cite{chandra_bayesian_2024}.

In Stage 3, we implement post-pruning strategies, where we remove unnecessary weights/biases, with the goal of obtaining compact BNNs without compromising performance accuracy. The weights are sorted according to a pruning criterion, such as a signal-to-noise ratio, which assesses the relative importance of each weight. The weights that fall below a user-defined threshold ($\lambda$) are set to zero, effectively removing them from the model hereafter. This pruning step reduces model complexity by eliminating low-impact weights, making the network compact and potentially more efficient. 

The pruned-BNN undergoes resampling in Stage 4, which repeats the Langevin MCMC sampling (training) process for the weights/biases that remained post-pruning, refining their posterior distribution.  The rationale is recapturing the lost information from the pruned weights into the remaining network neurons. This re-sampling strategy is novel to this application of STN and SPN as compared to \cite{graves_practical_2011, nalisnick_priors_2018}. Finally, in the fifth stage, we evaluate the pruned BNN model on the test data to measure its predictive accuracy and compare pre-pruning with post-pruning and post-pruning with resampling. This assessment ensures that the pruning process achieves a good balance between model efficiency and performance, resulting in a compact BNN that retains its predictive capabilities.

We need to check if the compact BNN converged better than standard BNNs. It is necessary to ensure that the MCMC chains have run long enough to converge to the target posterior distribution. Without proper convergence, estimates or inferences from the simulations may be unreliable or biased.   Therefore, we use the  Gelman-Rubin \cite{gelman_inference_1992} diagnostic, also known as the potential scale reduction factor (PSRF), to assess the convergence of the compact BNN. The diagnostic compares (\(\hat{R}\)) the variability between multiple MCMC chains, initialised from different starting points to the variability within each chain. When convergence is achieved, the between-chain and within-chain variances are expected to be similar.

\subsubsection{Datasets} 

We utilise benchmark datasets that encompass a diverse range of regression and classification tasks, spanning different domains and levels of complexity. Each dataset presents unique challenges, such as varying numbers of classes, features, and instances, which test the robustness and effectiveness of the pruning methods employed. 

We employ three datasets for the regression/prediction tasks including Lazer \cite{weigend_time_2018}, Sunspots \cite{solar_influences_data_analysis_center_sunspots_2018}, and Abalone (regression) \cite{warwick_nash_abalone_1994}.  In classification tasks, we employ the Ionosphere dataset, which is a binary classification task comprising 2 classes \cite{v_sigillito_ionosphere_1989}. We also use the Iris dataset \cite{r_a_fisher_iris_1936} which is prominent in machine learning and the Abalone dataset presented as a 4-class classification task \cite{warwick_nash_abalone_1994}. 
Our selected datasets span a broad spectrum of classification and regression problems, allowing for a thorough evaluation of the pruning methods across different domains and model complexities. The diversity in class structures, feature spaces, and data types makes these datasets ideal for testing the resilience of the pruning strategies under various conditions.

In addition to these benchmark datasets, we used two marine reef drill-core lithology classification datasets from the Integrated Ocean Drilling Program (IODP) Expedition 325 (Great Barrier Reef Environmental Changes) \cite{webster_proceedings_2011}  and Expedition 310 (Tahiti Sea Level) 310 \cite{camoin_proceedings_2007}. Expeditions 310 and 325 were international scientific efforts to understand past sea-level and climate changes and their impact on coral reef ecosystems. Expedition 310 \cite{camoin_proceedings_2007} )conducted in 2005), focused on Tahiti's coral reefs to reconstruct sea levels during the last deglaciation (around 20,000 to 10,000 years ago) by analysing fossil coral reef structure and composition. Similarly, Expedition 325 \cite{webster_proceedings_2011} (conducted in 2010), explored the Great Barrier Reef (GBR). The drilling of selected areas of the GBR has enabled the study of environmental changes over the past 30,000 years \cite{woodroffe_coral_2014, webster_response_2018}, particularly focusing on sea-level fluctuations, temperature changes, and their impact on reef growth \cite{sanborn_new_2017}. The reef core data has been used for analysis using machine learning and computer vision methods, and also for developing geoscientific models such as pyReef-Core which has been combined with Bayesian inference to estimate unknown parameters \cite{pall_bayesreef_2020}. Deo et al. \cite{deo_reefcoreseg_2024}  used this data to aid in the segmentation of drill core image data.

\begin{figure}[htb!]
    \centering
    \includegraphics[width=0.4\linewidth]{ 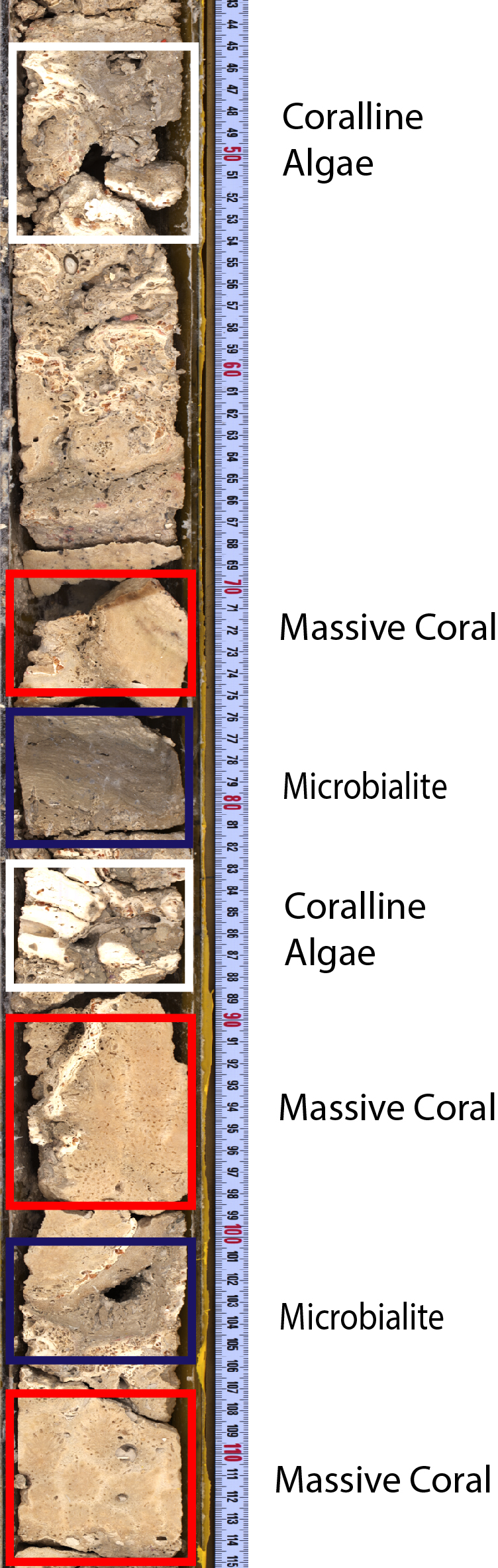}
    \caption{Lithology classification of drill core through visual analysis for a segment of Core 5R from Expedition 325 drill hole number M0033A. The core is taken at 43 meters depth below the seafloor.  }
    \label{fig:core_33A_lithology}
\end{figure}

\begin{figure}[htb!]
    \centering
    \includegraphics[width=0.9\linewidth]{ 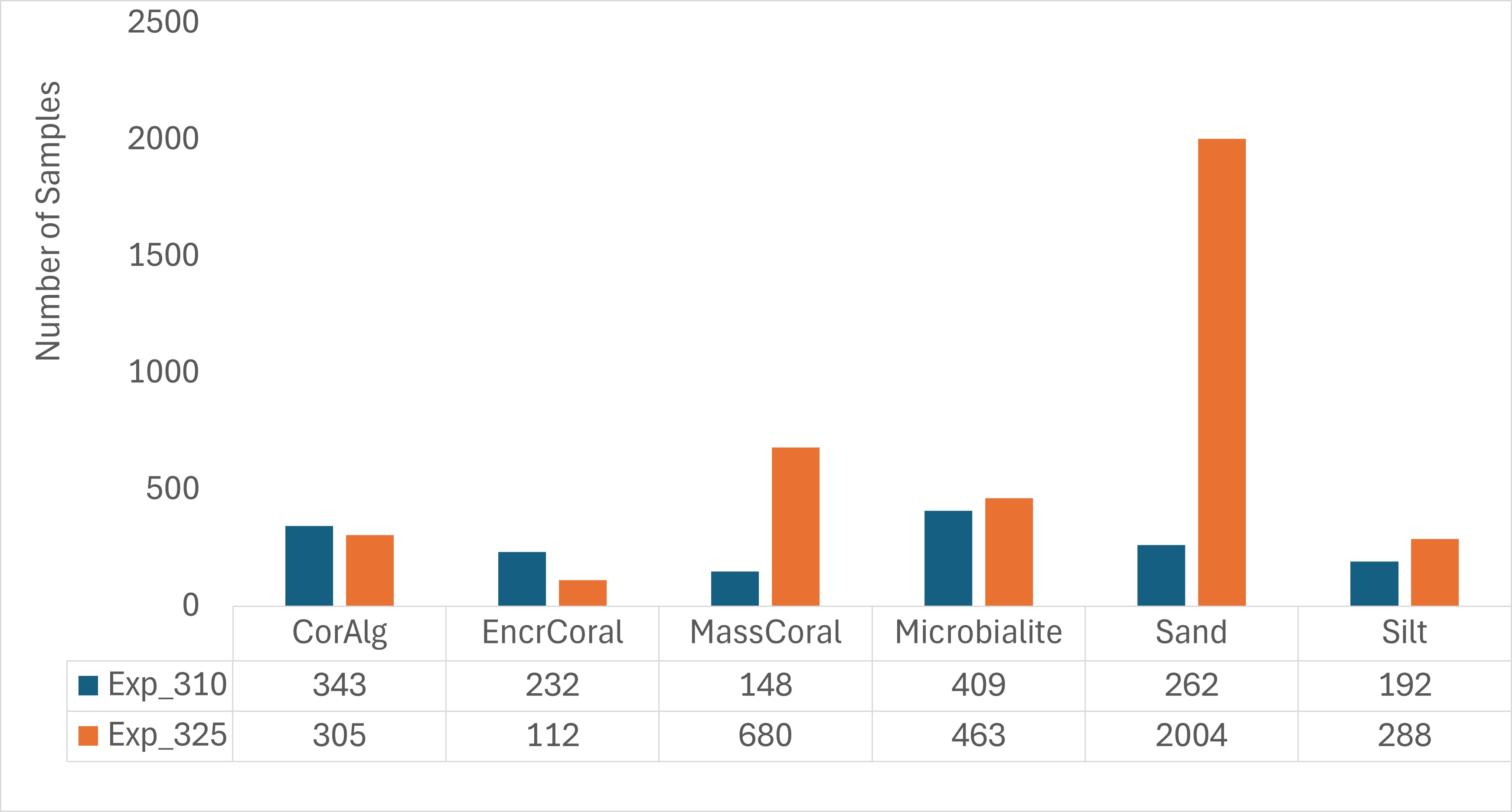}
    \caption{Class distribution for Expedition 325 and 310 Datasets}
    \label{fig:exp325_310_data}
\end{figure}

The reef-core (drilled) datasets represent fossil coral reef lithology classification tasks with six distinct target classes on drill cores extracted offshore on these IODP expeditions.  These datasets were curated by mapping physical properties measurements taken on reef drill cores measured using a \textit{multi-sensor core logger} (MSCL) to physical lithologies seen through expert visual analysis of the reef cores. Figure \ref{fig:core_33A_lithology}  presents a sample drill core section taken from Expedition 325 that has been classified into 3 distinct lithologies.   The data comprises 3 features: bulk density, porosity, and resistivity.  The physical properties were classified into 6 distinct lithologies (massive coral, encrusted coral, coralline algae, microbialite, sand, and silt). Figure \ref{fig:exp325_310_data} presents the relative abundance of each lithology in the two datasets.  Each expedition collected samples across the same categories, but Expedition 325 generally collected more samples, especially in the Sand and Massive Coral categories. In Expedition 325, the 'sand' stands out with a significantly higher sample count (2004 samples) compared to Expedition 310 (262 samples), while 'massive coral' also shows a higher count in Expedition 325 (680 samples) than in Expedition 310 (148 samples). The other categories show smaller differences between the two expeditions. Expedition 310 collected a more balanced set of samples across all categories without any extreme outliers, likely due to significantly higher recovery rates of samples during the drilling process\cite{camoin_proceedings_2007}. 

\section{Experiments and Results}
\label{sec:results}

\subsection{Data Preprocessing and model selection}

We preprocessed each dataset by normalising all input features to ensure consistency across models. We utilised all features present in each dataset as shown in Table \ref{tab:models} with a standard 60:40 train-test split to each dataset for model training and evaluation.   As BNN models have challenges in convergence \cite{chandra_bayesian_2024}, we run multiple independent experiments to show that pruning has a similar effect on the model even if there are convergence problems. For each dataset, we sampled each BNN model 50,000 times and then resampled post-pruning for an additional 1000 samples without burn-in.  We execute 30 independent BNN model training runs to capture a comprehensive range of performance metrics and report the mean and standard deviation.


\begin{table*}[htb!]
\footnotesize
\centering

\begin{tabular}{|c|c|c|c|c|c|c|}
\hline
\textbf{Dataset} &
  \textbf{Train Size} &
  \textbf{Test Size} &
  \textbf{Num. Features} &
  \textbf{\begin{tabular}[c]{@{}c@{}}Network Structure \\ {[}Input,   Hidden, Output{]}\end{tabular}} &
  \textbf{\begin{tabular}[c]{@{}c@{}}Number\\ Parameters\end{tabular}} &
  \textbf{\begin{tabular}[c]{@{}c@{}}Sampling time \\ {[}mins/run{]}\end{tabular}} \\ \hline
\textbf{Ionosphere (2 class)} & 211  & 140  & 34 & 34,   50, 2 & 1852 & 80  \\ \hline
\textbf{Iris   (3 class)}     & 90   & 60   & 4  & 4, 12, 3    & 100  & 13  \\ \hline
\textbf{Abalone (4 class)}    & 2506 & 1670 & 8  & 8,   12, 4  & 144  & 570 \\ \hline
\textbf{Exp   325 (6 class)}  & 2311 & 1541 & 3  & 3, 8, 6     & 72   & 380 \\ \hline
\textbf{Exp 310 (6 class)}    & 952  & 634  & 3  & 3,   8, 6   & 72   & 200 \\ \hline
\textbf{Lazer}                & 299  & 199  & 4  & 4, 5, 1     & 32   & 13  \\ \hline
\textbf{Sunspots}             & 817  & 544  & 4  & 4,   5, 1   & 32   & 44  \\ \hline
\textbf{Abalone}              & 2506 & 1670 & 8  & 8, 12, 1    & 108  & 700 \\ \hline
\end{tabular}
\caption{Datasets and BNN model configuration including the number of training and testing samples, the number of input features, BNN architecture (input, hidden, and output neurons), the total number of parameters, and the average time  (in minutes) for  BNN sampling (training).}
\label{tab:models}

\end{table*}

\subsection{Evaluation Metrics}

In this study, we use three evaluation metrics to assess model performance, depending on the task associated with each dataset. 
\begin{enumerate}

   \item The  Root Mean Squared Error (RMSE) measures the average squared difference between the actual and predicted values for regression tasks. A lower RMSE  indicates better performance, while an increasing RMSE suggests overfitting or poor model generalisation.

   \item The classification accuracy measures the proportion of correct predictions made by the model out of the total predictions. It is a common metric in classification tasks to evaluate how well the model is distinguishing between different classes. Higher accuracy indicates better performance in terms of correctly classifying instances.

   \item  The Receiver Operating Characteristic (ROC) curve \cite{bradley_use_1997}  evaluates the performance of the models, for classification problems typically used for binary classification but can be extended to multi-class classification using a one-vs-all approach \cite{rocha_multiclass_2013}. The Area-Under-Curve (AUC) \cite{huang_using_2005} represents the area under the ROC curve, which is a measure prominently used for the accuracy of predictions.   

\end{enumerate}
\subsection{Results and Analysis}

\subsection{Benchamrk datasets}

\begin{table}[htb!]
\footnotesize
\centering

\setlength\tabcolsep{0pt}
\begin{tabular*}{\linewidth}{@{\extracolsep{\fill}}|c|cccccc|}

\hline
\textbf{Pruning   Level} & \multicolumn{6}{c|}{\textbf{0.25}} \\ \hline
\textbf{Pruning Method} & \multicolumn{2}{c|}{\textbf {RND}} & \multicolumn{2}{c|}{\textbf {SPN}} & \multicolumn{2}{c|}{\textbf {STN}} \\ \hline
\textbf{Resamping} & \multicolumn{1}{c|}{\textbf{No}} & \multicolumn{1}{c|}{\textbf{Yes}} & \multicolumn{1}{c|}{\textbf{No}} & \multicolumn{1}{c|}{\textbf{Yes}} & \multicolumn{1}{c|}{\textbf{No}} & \textbf{Yes} \\ \hline
\multirow{2}{*}{\textbf{Sunspots}} & \multicolumn{1}{c|}{0.090} & \multicolumn{1}{c|}{0.065} & \multicolumn{1}{c|}{0.082} & \multicolumn{1}{c|}{0.063} & \multicolumn{1}{c|}{0.082} & 0.062 \\ \cline{2-7} 
 & \multicolumn{1}{c|}{±0.012} & \multicolumn{1}{c|}{±0.001} & \multicolumn{1}{c|}{±0.034} & \multicolumn{1}{c|}{±0.002} & \multicolumn{1}{c|}{±0.025} & ±0.001 \\ \hline
\multirow{2}{*}{\textbf{Lazer}} & \multicolumn{1}{c|}{0.054} & \multicolumn{1}{c|}{0.027} & \multicolumn{1}{c|}{0.081} & \multicolumn{1}{c|}{0.025} & \multicolumn{1}{c|}{0.061} & 0.022 \\ \cline{2-7} 
 & \multicolumn{1}{c|}{±0.026} & \multicolumn{1}{c|}{±0.004} & \multicolumn{1}{c|}{±0.059} & \multicolumn{1}{c|}{±0.005} & \multicolumn{1}{c|}{±0.046} & ±0.002 \\ \hline
\multirow{2}{*}{\textbf{Abalone}} & \multicolumn{1}{c|}{0.137} & \multicolumn{1}{c|}{0.079} & \multicolumn{1}{c|}{0.105} & \multicolumn{1}{c|}{0.078} & \multicolumn{1}{c|}{0.102} & 0.078 \\ \cline{2-7} 
 & \multicolumn{1}{c|}{±0.049} & \multicolumn{1}{c|}{±0.001} & \multicolumn{1}{c|}{±0.031} & \multicolumn{1}{c|}{±0.000} & \multicolumn{1}{c|}{±0.019} & ±0.000 \\ \hline
\textbf{Pruning   Level} & \multicolumn{6}{c|}{\textbf{0.5}} \\ \hline
\textbf{Pruning   Method} & \multicolumn{2}{c|}{\textbf {RND}} & \multicolumn{2}{c|}{\textbf {SPN}} & \multicolumn{2}{c|}{\textbf {STN}} \\ \hline
\textbf{Resamping} & \multicolumn{1}{c|}{\textbf{No}} & \multicolumn{1}{c|}{\textbf{Yes}} & \multicolumn{1}{c|}{\textbf{No}} & \multicolumn{1}{c|}{\textbf{Yes}} & \multicolumn{1}{c|}{\textbf{No}} & \textbf{Yes} \\ \hline
\multirow{2}{*}{\textbf{Sunspots}} & \multicolumn{1}{c|}{0.162} & \multicolumn{1}{c|}{0.069} & \multicolumn{1}{c|}{0.149} & \multicolumn{1}{c|}{0.067} & \multicolumn{1}{c|}{0.143} & 0.065 \\ \cline{2-7} 
 & \multicolumn{1}{c|}{±0.045} & \multicolumn{1}{c|}{±0.002} & \multicolumn{1}{c|}{±0.086} & \multicolumn{1}{c|}{±0.004} & \multicolumn{1}{c|}{±0.080} & ±0.002 \\ \hline
\multirow{2}{*}{\textbf{Lazer}} & \multicolumn{1}{c|}{0.427} & \multicolumn{1}{c|}{0.157} & \multicolumn{1}{c|}{0.133} & \multicolumn{1}{c|}{0.037} & \multicolumn{1}{c|}{0.157} & 0.033 \\ \cline{2-7} 
 & \multicolumn{1}{c|}{±0.137} & \multicolumn{1}{c|}{±0.003} & \multicolumn{1}{c|}{±0.071} & \multicolumn{1}{c|}{±0.013} & \multicolumn{1}{c|}{±0.141} & ±0.009 \\ \hline
\multirow{2}{*}{\textbf{Abalone}} & \multicolumn{1}{c|}{0.225} & \multicolumn{1}{c|}{0.081} & \multicolumn{1}{c|}{0.183} & \multicolumn{1}{c|}{0.079} & \multicolumn{1}{c|}{0.163} & 0.079 \\ \cline{2-7} 
 & \multicolumn{1}{c|}{±0.028} & \multicolumn{1}{c|}{±0.001} & \multicolumn{1}{c|}{±0.106} & \multicolumn{1}{c|}{±0.001} & \multicolumn{1}{c|}{±0.064} & ±0.000 \\ \hline
\textbf{Pruning   Level} & \multicolumn{6}{c|}{\textbf{0.75}} \\ \hline
\textbf{Pruning   Method} & \multicolumn{2}{c|}{\textbf {RND}} & \multicolumn{2}{c|}{\textbf {SPN}} & \multicolumn{2}{c|}{\textbf {STN}} \\ \hline
\textbf{Resamping} & \multicolumn{1}{c|}{\textbf{No}} & \multicolumn{1}{c|}{\textbf{Yes}} & \multicolumn{1}{c|}{\textbf{No}} & \multicolumn{1}{c|}{\textbf{Yes}} & \multicolumn{1}{c|}{\textbf{No}} & \textbf{Yes} \\ \hline
\multirow{2}{*}{\textbf{Sunspots}} & \multicolumn{1}{c|}{0.298} & \multicolumn{1}{c|}{0.167} & \multicolumn{1}{c|}{0.189} & \multicolumn{1}{c|}{0.079} & \multicolumn{1}{c|}{0.225} & 0.084 \\ \cline{2-7} 
 & \multicolumn{1}{c|}{±0.155} & \multicolumn{1}{c|}{±0.012} & \multicolumn{1}{c|}{±0.114} & \multicolumn{1}{c|}{±0.018} & \multicolumn{1}{c|}{±0.164} & ±0.035 \\ \hline
\multirow{2}{*}{\textbf{Lazer}} & \multicolumn{1}{c|}{0.464} & \multicolumn{1}{c|}{0.231} & \multicolumn{1}{c|}{0.319} & \multicolumn{1}{c|}{0.106} & \multicolumn{1}{c|}{0.381} & 0.149 \\ \cline{2-7} 
 & \multicolumn{1}{c|}{±0.217} & \multicolumn{1}{c|}{±0.129} & \multicolumn{1}{c|}{±0.193} & \multicolumn{1}{c|}{±0.068} & \multicolumn{1}{c|}{±0.209} & ±0.147 \\ \hline
\multirow{2}{*}{\textbf{Abalone}} & \multicolumn{1}{c|}{0.171} & \multicolumn{1}{c|}{0.086} & \multicolumn{1}{c|}{0.234} & \multicolumn{1}{c|}{0.080} & \multicolumn{1}{c|}{0.268} & 0.081 \\ \cline{2-7} 
 & \multicolumn{1}{c|}{±0.039} & \multicolumn{1}{c|}{±0.001} & \multicolumn{1}{c|}{±0.108} & \multicolumn{1}{c|}{±0.001} & \multicolumn{1}{c|}{±0.131} & ±0.002 \\ \hline
\end{tabular*}

\caption{Performance accuracy (RMSE) given by mean and standard deviation (mean ± std) for Sunspots, Lazer, and Abalone datasets at different pruning levels. The table compares random pruning (RND), signal-plus-noise (SPN), and signal-to-noise (STN) with and without resampling. 
}
\label{tab:regression}
\end{table}

\begin{figure}[htb!]
    \centering
    \includegraphics[width=\linewidth]{ 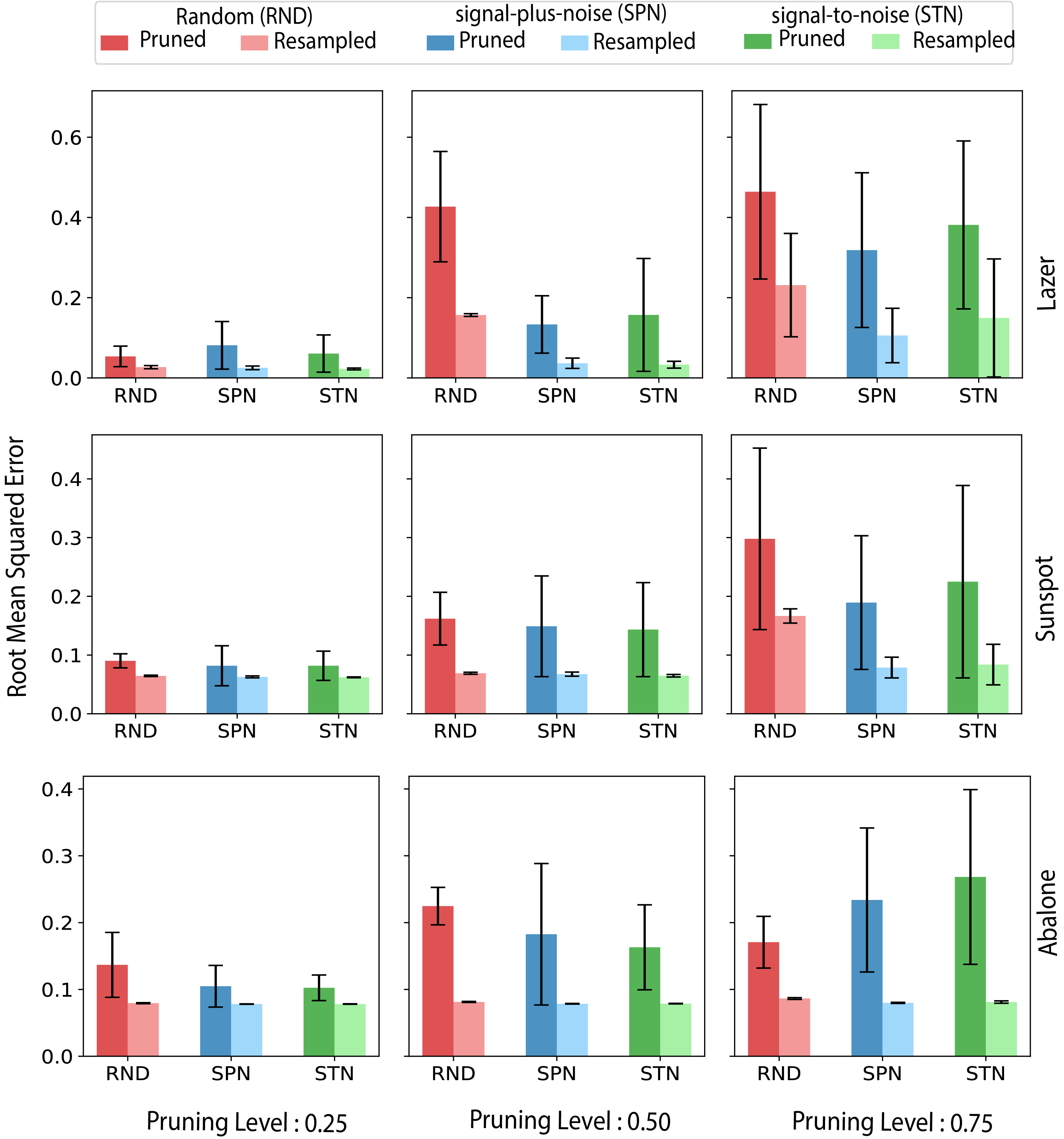}
    \caption{Performance accuracy (RMSE) for different pruning methods and pruning levels for given datasets (Lazer, Sunspot, and Abalone). Each method is distinguished by a unique colour scheme, with darker shades representing the original network and lighter shades representing resampled networks. Error bars indicate the standard deviation, highlighting variability in RMSE. }
    
    
    \label{fig:regression_resample}
\end{figure}
We begin by evaluating the effect of the different pruning methods, including signal-plus-noise (SPN), signal-to-noise (STN), and random (RND) pruning for the different datasets.   Figure \ref{fig:regression_resample} evaluates the selected pruning methods for BNNs using three regression datasets: Lazer, Sunspot, and Abalone. We observe that the accuracy  (RMSE) generally deteriorates as the pruning level rises, particularly for random pruning. This trend is evident across all datasets, reflecting the challenges of maintaining network performance when a significant proportion of parameters are removed randomly. We observe that the performance accuracy varies across datasets (Figure \ref{fig:regression_resample}). For the Abalone dataset, the RMSE remains relatively stable for all methods and pruning methods and levels.  All the models can capture the intricacies of the Abalone dataset when resampled. In the Sunspot dataset, we see the impact of the structured pruning methods over random pruning at higher pruning levels.  We can clearly see that when a large number (75\%) of parameters are randomly removed, we cannot recapture the model performance with resampling.   In the Lazer dataset, we continue to see the significance of structured pruning as seen in the Sunspot dataset.  With high pruning rates, structured pruning outperforms random pruning. Structured pruning methods (SPN and STN) can recapture lost knowledge even at high pruning rates, indicating better preservation of critical parameters. At the highest pruning level (0.75), the  RND pruning method consistently shows the worst accuracy, highlighting its sensitivity to high pruning levels.

Resampling has a notable impact on reducing RMSE across all methods and pruning levels, as shown in Figure \ref{fig:regression_resample}. We observe that resampled BNNs (lighter shades) consistently outperform their original counterparts (darker shades), especially for the  RND method. This suggests that resampling effectively compensates for the randomness introduced by unstructured pruning, leading to more stable and accurate predictions. For structured pruning methods (STN and SPN), the improvement with resampling is less pronounced but still noticeable.  We find that in these regression datasets, pruning with SPN is able to get a more precise prediction than STN at larger pruning rates.  This is consistent in all datasets. However, it can be clearly seen in the Lazer dataset.  Table \ref{tab:regression} shows that the standard deviations on the resampled models of SPN are lower than those of STN models.   This indicates that SPN has more consistent predictive performance for regression tasks.


\begin{table}[htb!]
\footnotesize
\centering
\setlength\tabcolsep{0pt}
\begin{tabular*}{\linewidth}{@{\extracolsep{\fill}}|c|cccccc|}
\hline
\textbf{Pruning Level} & \multicolumn{6}{c|}{\textbf{0.25}} \\ \hline
\textbf{Pruning   Method} & \multicolumn{2}{c|}{\textbf {RND}} & \multicolumn{2}{c|}{\textbf {SPN}} & \multicolumn{2}{c|}{\textbf {STN}} \\ \hline
\textbf{Resamping} & \multicolumn{1}{c|}{\textbf{No}} & \multicolumn{1}{c|}{\textbf{Yes}} & \multicolumn{1}{c|}{\textbf{No}} & \multicolumn{1}{c|}{\textbf{Yes}} & \multicolumn{1}{c|}{\textbf{No}} & \textbf{Yes} \\ \hline
\multirow{2}{*}{\textbf{Ionosphere (2 class)}} & \multicolumn{1}{c|}{40.00} & \multicolumn{1}{c|}{68.09} & \multicolumn{1}{c|}{76.70} & \multicolumn{1}{c|}{92.73} & \multicolumn{1}{c|}{77.48} & 92.55 \\ \cline{2-7} 
 & \multicolumn{1}{c|}{±1.013} & \multicolumn{1}{c|}{±23.780} & \multicolumn{1}{c|}{±5.147} & \multicolumn{1}{c|}{±2.229} & \multicolumn{1}{c|}{±4.630} & ±1.482 \\ \hline
\multirow{2}{*}{\textbf{Iris (3 class)}} & \multicolumn{1}{c|}{91.26} & \multicolumn{1}{c|}{96.23} & \multicolumn{1}{c|}{96.87} & \multicolumn{1}{c|}{97.54} & \multicolumn{1}{c|}{97.72} & 98.03 \\ \cline{2-7} 
 & \multicolumn{1}{c|}{±10.785} & \multicolumn{1}{c|}{±1.037} & \multicolumn{1}{c|}{±1.068} & \multicolumn{1}{c|}{±1.018} & \multicolumn{1}{c|}{±0.769} & ±0.747 \\ \hline
\multirow{2}{*}{\textbf{Abalone (4 class)}} & \multicolumn{1}{c|}{77.12} & \multicolumn{1}{c|}{77.90} & \multicolumn{1}{c|}{77.33} & \multicolumn{1}{c|}{78.42} & \multicolumn{1}{c|}{78.29} & 78.26 \\ \cline{2-7} 
 & \multicolumn{1}{c|}{±1.376} & \multicolumn{1}{c|}{±2.080} & \multicolumn{1}{c|}{±1.482} & \multicolumn{1}{c|}{±2.152} & \multicolumn{1}{c|}{±2.124} & ±2.051 \\ \hline
\multirow{2}{*}{\textbf{Exp 325 (6 class)}} & \multicolumn{1}{c|}{41.47} & \multicolumn{1}{c|}{53.24} & \multicolumn{1}{c|}{56.40} & \multicolumn{1}{c|}{58.77} & \multicolumn{1}{c|}{58.89} & 58.91 \\ \cline{2-7} 
 & \multicolumn{1}{c|}{±16.820} & \multicolumn{1}{c|}{±3.529} & \multicolumn{1}{c|}{±9.425} & \multicolumn{1}{c|}{±4.726} & \multicolumn{1}{c|}{±4.511} & ±4.426 \\ \hline
\multirow{2}{*}{\textbf{Exp 310 (6 class)}} & \multicolumn{1}{c|}{21.41} & \multicolumn{1}{c|}{27.13} & \multicolumn{1}{c|}{36.89} & \multicolumn{1}{c|}{37.11} & \multicolumn{1}{c|}{36.33} & 36.78 \\ \cline{2-7} 
 & \multicolumn{1}{c|}{±7.903} & \multicolumn{1}{c|}{±3.153} & \multicolumn{1}{c|}{±4.183} & \multicolumn{1}{c|}{±4.389} & \multicolumn{1}{c|}{±4.498} & ±4.444 \\ \hline
\textbf{Pruning   Level} & \multicolumn{6}{c|}{\textbf{0.5}} \\ \hline
\textbf{Pruning   Method} & \multicolumn{2}{c|}{\textbf {RND}} & \multicolumn{2}{c|}{\textbf {SPN}} & \multicolumn{2}{c|}{\textbf {STN}} \\ \hline
\textbf{Resamping} & \multicolumn{1}{c|}{\textbf{No}} & \multicolumn{1}{c|}{\textbf{Yes}} & \multicolumn{1}{c|}{\textbf{No}} & \multicolumn{1}{c|}{\textbf{Yes}} & \multicolumn{1}{c|}{\textbf{No}} & \textbf{Yes} \\ \hline
\multirow{2}{*}{\textbf{Ionosphere (2 class)}} & \multicolumn{1}{c|}{40.18} & \multicolumn{1}{c|}{54.18} & \multicolumn{1}{c|}{59.26} & \multicolumn{1}{c|}{90.35} & \multicolumn{1}{c|}{61.21} & 90.25 \\ \cline{2-7} 
 & \multicolumn{1}{c|}{±0.576} & \multicolumn{1}{c|}{±19.429} & \multicolumn{1}{c|}{±15.597} & \multicolumn{1}{c|}{±3.183} & \multicolumn{1}{c|}{±15.333} & ±2.622 \\ \hline
\multirow{2}{*}{\textbf{Iris (3 class)}} & \multicolumn{1}{c|}{38.46} & \multicolumn{1}{c|}{37.44} & \multicolumn{1}{c|}{91.62} & \multicolumn{1}{c|}{95.54} & \multicolumn{1}{c|}{93.64} & 97.18 \\ \cline{2-7} 
 & \multicolumn{1}{c|}{±0.000} & \multicolumn{1}{c|}{±3.903} & \multicolumn{1}{c|}{±5.056} & \multicolumn{1}{c|}{±1.899} & \multicolumn{1}{c|}{±5.062} & ±1.330 \\ \hline
\multirow{2}{*}{\textbf{Abalone (4 class)}} & \multicolumn{1}{c|}{64.84} & \multicolumn{1}{c|}{76.77} & \multicolumn{1}{c|}{76.51} & \multicolumn{1}{c|}{78.37} & \multicolumn{1}{c|}{77.64} & 78.27 \\ \cline{2-7} 
 & \multicolumn{1}{c|}{±23.712} & \multicolumn{1}{c|}{±0.874} & \multicolumn{1}{c|}{±0.303} & \multicolumn{1}{c|}{±2.118} & \multicolumn{1}{c|}{±1.830} & ±2.164 \\ \hline
\multirow{2}{*}{\textbf{Exp 325 (6 class)}} & \multicolumn{1}{c|}{34.72} & \multicolumn{1}{c|}{47.35} & \multicolumn{1}{c|}{53.70} & \multicolumn{1}{c|}{56.87} & \multicolumn{1}{c|}{54.05} & 58.54 \\ \cline{2-7} 
 & \multicolumn{1}{c|}{±17.625} & \multicolumn{1}{c|}{±10.592} & \multicolumn{1}{c|}{±12.900} & \multicolumn{1}{c|}{±7.789} & \multicolumn{1}{c|}{±11.927} & ±4.356 \\ \hline
\multirow{2}{*}{\textbf{Exp 310 (6 class)}} & \multicolumn{1}{c|}{20.34} & \multicolumn{1}{c|}{24.76} & \multicolumn{1}{c|}{34.33} & \multicolumn{1}{c|}{35.61} & \multicolumn{1}{c|}{35.02} & 35.82 \\ \cline{2-7} 
 & \multicolumn{1}{c|}{±7.308} & \multicolumn{1}{c|}{±6.226} & \multicolumn{1}{c|}{±5.113} & \multicolumn{1}{c|}{±4.779} & \multicolumn{1}{c|}{±4.610} & ±4.475 \\ \hline
\textbf{Pruning   Level} & \multicolumn{6}{c|}{\textbf{0.75}} \\ \hline
\textbf{Pruning   Method} & \multicolumn{2}{c|}{\textbf {RND}} & \multicolumn{2}{c|}{\textbf {SPN}} & \multicolumn{2}{c|}{\textbf {STN}} \\ \hline
\textbf{Resamping} & \multicolumn{1}{c|}{\textbf{No}} & \multicolumn{1}{c|}{\textbf{Yes}} & \multicolumn{1}{c|}{\textbf{No}} & \multicolumn{1}{c|}{\textbf{Yes}} & \multicolumn{1}{c|}{\textbf{No}} & \textbf{Yes} \\ \hline
\multirow{2}{*}{\textbf{Ionosphere (2 class)}} & \multicolumn{1}{c|}{42.38} & \multicolumn{1}{c|}{56.77} & \multicolumn{1}{c|}{50.32} & \multicolumn{1}{c|}{78.58} & \multicolumn{1}{c|}{50.92} & 77.38 \\ \cline{2-7} 
 & \multicolumn{1}{c|}{±7.438} & \multicolumn{1}{c|}{±20.923} & \multicolumn{1}{c|}{±15.999} & \multicolumn{1}{c|}{±18.998} & \multicolumn{1}{c|}{±14.450} & ±18.517 \\ \hline
\multirow{2}{*}{\textbf{Iris (3 class)}} & \multicolumn{1}{c|}{34.36} & \multicolumn{1}{c|}{32.82} & \multicolumn{1}{c|}{68.95} & \multicolumn{1}{c|}{79.15} & \multicolumn{1}{c|}{69.31} & 84.82 \\ \cline{2-7} 
 & \multicolumn{1}{c|}{±6.920} & \multicolumn{1}{c|}{±7.541} & \multicolumn{1}{c|}{±17.780} & \multicolumn{1}{c|}{±15.509} & \multicolumn{1}{c|}{±17.863} & ±11.415 \\ \hline
\multirow{2}{*}{\textbf{Abalone (4 class)}} & \multicolumn{1}{c|}{54.72} & \multicolumn{1}{c|}{74.44} & \multicolumn{1}{c|}{74.38} & \multicolumn{1}{c|}{78.12} & \multicolumn{1}{c|}{74.81} & 78.55 \\ \cline{2-7} 
 & \multicolumn{1}{c|}{±31.218} & \multicolumn{1}{c|}{±11.940} & \multicolumn{1}{c|}{±9.740} & \multicolumn{1}{c|}{±2.120} & \multicolumn{1}{c|}{±10.411} & ±2.228 \\ \hline
\multirow{2}{*}{\textbf{Exp 325 (6 class)}} & \multicolumn{1}{c|}{35.18} & \multicolumn{1}{c|}{49.71} & \multicolumn{1}{c|}{43.89} & \multicolumn{1}{c|}{54.38} & \multicolumn{1}{c|}{42.18} & 56.05 \\ \cline{2-7} 
 & \multicolumn{1}{c|}{±18.305} & \multicolumn{1}{c|}{±6.031} & \multicolumn{1}{c|}{±20.606} & \multicolumn{1}{c|}{±11.142} & \multicolumn{1}{c|}{±20.492} & ±8.226 \\ \hline
\multirow{2}{*}{\textbf{Exp 310 (6 class)}} & \multicolumn{1}{c|}{19.48} & \multicolumn{1}{c|}{23.83} & \multicolumn{1}{c|}{27.89} & \multicolumn{1}{c|}{28.78} & \multicolumn{1}{c|}{27.28} & 31.45 \\ \cline{2-7} 
 & \multicolumn{1}{c|}{±7.050} & \multicolumn{1}{c|}{±6.751} & \multicolumn{1}{c|}{±8.139} & \multicolumn{1}{c|}{±8.174} & \multicolumn{1}{c|}{±7.248} & ±8.512 \\ \hline
\end{tabular*}
\caption{Classification performance for various datasets at different pruning levels, comparing  RND, SPN, STN  pruning methods with and without resampling. We report the mean and standard deviation (std) classification accuracy for 30 independent model training  runs.}
\label{tab:classification}
\end{table}

\begin{figure}[htb!]
    \centering
    \includegraphics[width=\linewidth]{ 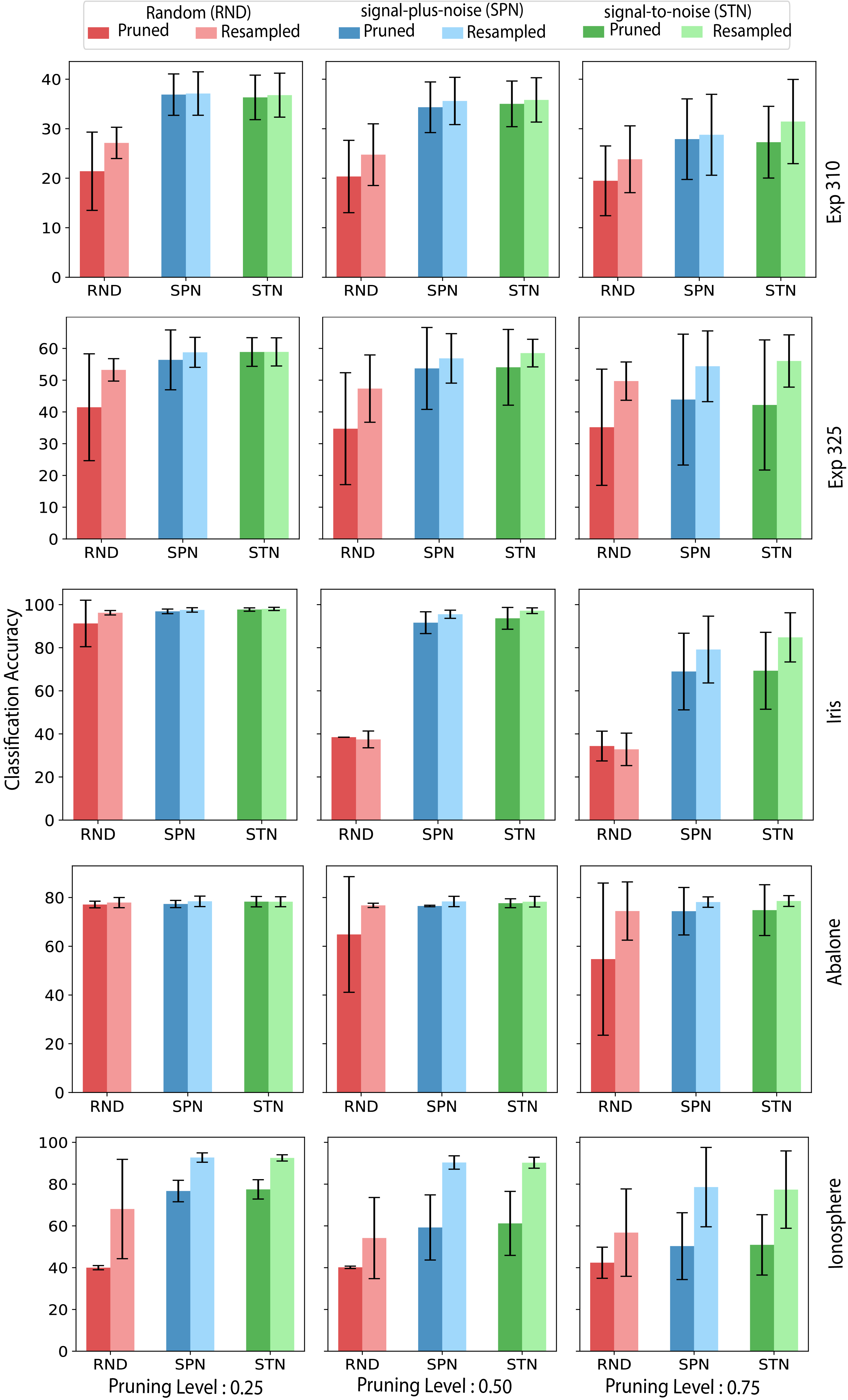}
    \caption{Classification accuracy of Bayesian neural networks across different pruning methods and pruning levels. Each method is represented by a unique colour scheme, with darker shades indicating the original network and lighter shades indicating resampled networks. The error bars represent the standard deviation, highlighting variability in performance. }
    \label{fig:classification_resample}
\end{figure}

Figure \ref{fig:classification_resample} presents the BNN model classification results, demonstrating that increasing the pruning level generally leads to a slight decline in classification accuracy. However, the severity of the impact varies across methods and datasets.  RND exhibits the most pronounced drop in accuracy, particularly at higher pruning levels (0.75), indicating its inability to preserve critical information. In contrast, SPN and STN consistently maintain higher accuracy, even under severe pruning, highlighting their robustness in retaining essential network parameters.  This is consistent with the regression performance.

There are two significant differences in the classification and regression model performance.  Firstly, we see that the performance of the models on Exp 310 and Exp 325 datasets are lower than those of the benchmark datasets.  We attribute this to the increased complexity of the datasets.  These datasets also have the highest number of classes being predicted.  These datasets also have half the number of input features for the number of predicted classes (3 features to 6 classes).  The second significant difference we see is that unlike in regression, STN consistently outperforms both RND and SPN across all classification datasets and pruning levels. The signal-to-noise ratio criterion effectively identifies and preserves critical weights, ensuring minimal degradation in model performance. Table \ref{tab:classification} also shows that STN has higher classification accuracy and lower standard deviations in the model performance at 75\% pruning as compared to SPN. RND consistently underperforms and exhibits higher variability, as evidenced by larger error bars, making it less suitable for high-stakes applications.  The inclusion of resampling significantly improves classification accuracy for all pruning methods. We can observe that the resampled BNNs (lighter shades in Figure \ref{fig:classification_resample}) consistently achieve higher classification accuracy than their original counterparts (darker shades). The improvement is most notable in the  RND method, where resampling effectively mitigates its inherent randomness and variability. 

\subsection{Convergence diagnostic }

\begin{figure*}[htb!]
    \centering
    \includegraphics[width=1\linewidth]{ 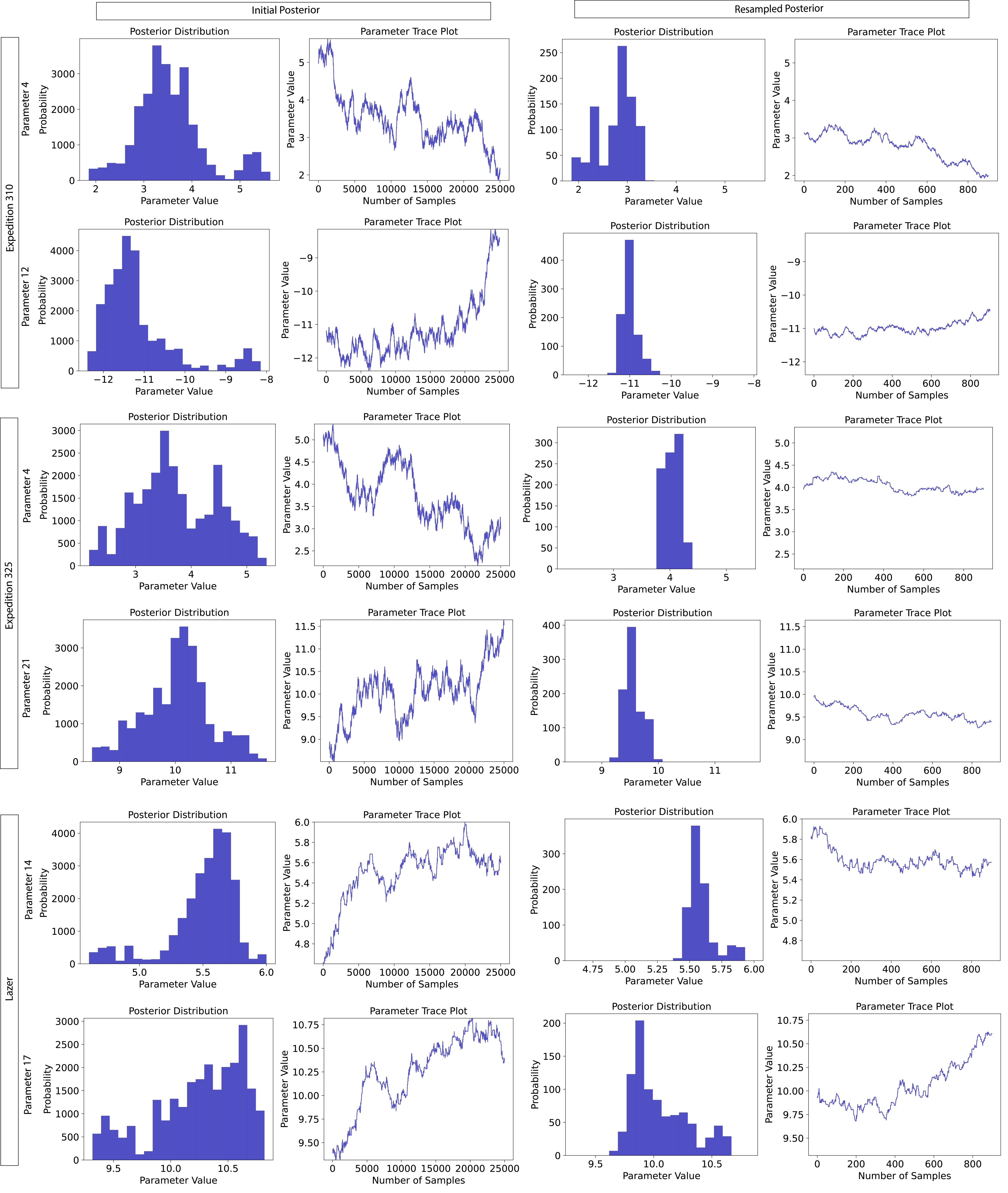}
    \caption{Posterior distribution and parameter trace plots for  BNNs models across different datasets and resampling scenarios. The left panels illustrate the initial posterior distributions and corresponding parameter trace plots for unpruned BNN models, showcasing parameter variability and convergence over 25,000 post-burn-in samples. The right panels depict the posterior distributions and trace plots after resampling for obtaining pruned BNN models; each row corresponds to a specific parameter from the dataset Expedition 310, Expedition 325, and Lazer.}
    \label{fig:traceplot}
\end{figure*}


Figure \ref{fig:traceplot} presents posterior distributions and trace plots for BNN model parameters for selected datasets, highlighting the effects of pruning and resampling. Each row corresponds to a randomly selected parameter from the posterior of the BNN model trained on datasets Expedition 310, Expedition 325, and Lazer.  The left panels represent the initial posterior distributions and trace plots, and the right panels depict the same metrics after pruning and resampling. The initial posterior distributions, observed in the left panels, are relatively wide, reflecting significant variability in parameter estimates. This variability is further corroborated by the trace plots, which show notable fluctuations in parameter values over 25,000 samples. The wide distributions and noisy trace plot regions indicate that the unpruned models are less confident in their parameter estimates, particularly for datasets such as Expedition 310 and Lazer. This suggests that these datasets may contain significant noise or redundant features, challenging the convergence and stability of the MCMC sampling process. The right panels illustrate the posterior distributions and trace plots following pruning and resampling for an additional 900 post-burnin samples.  A notable improvement is observed across all datasets, with posterior distributions becoming narrower and more concentrated around specific parameter values. This indicates that resampling post-pruning eliminates noisy or redundant parameters, enabling the model to focus on the most relevant features. The trace plots for resampled models demonstrate smoother convergence, with significantly reduced fluctuations in parameter values, further underscoring the stabilising effect of pruning.

\begin{figure}[htb!]
    \centering
    \includegraphics[width=1\linewidth]{ 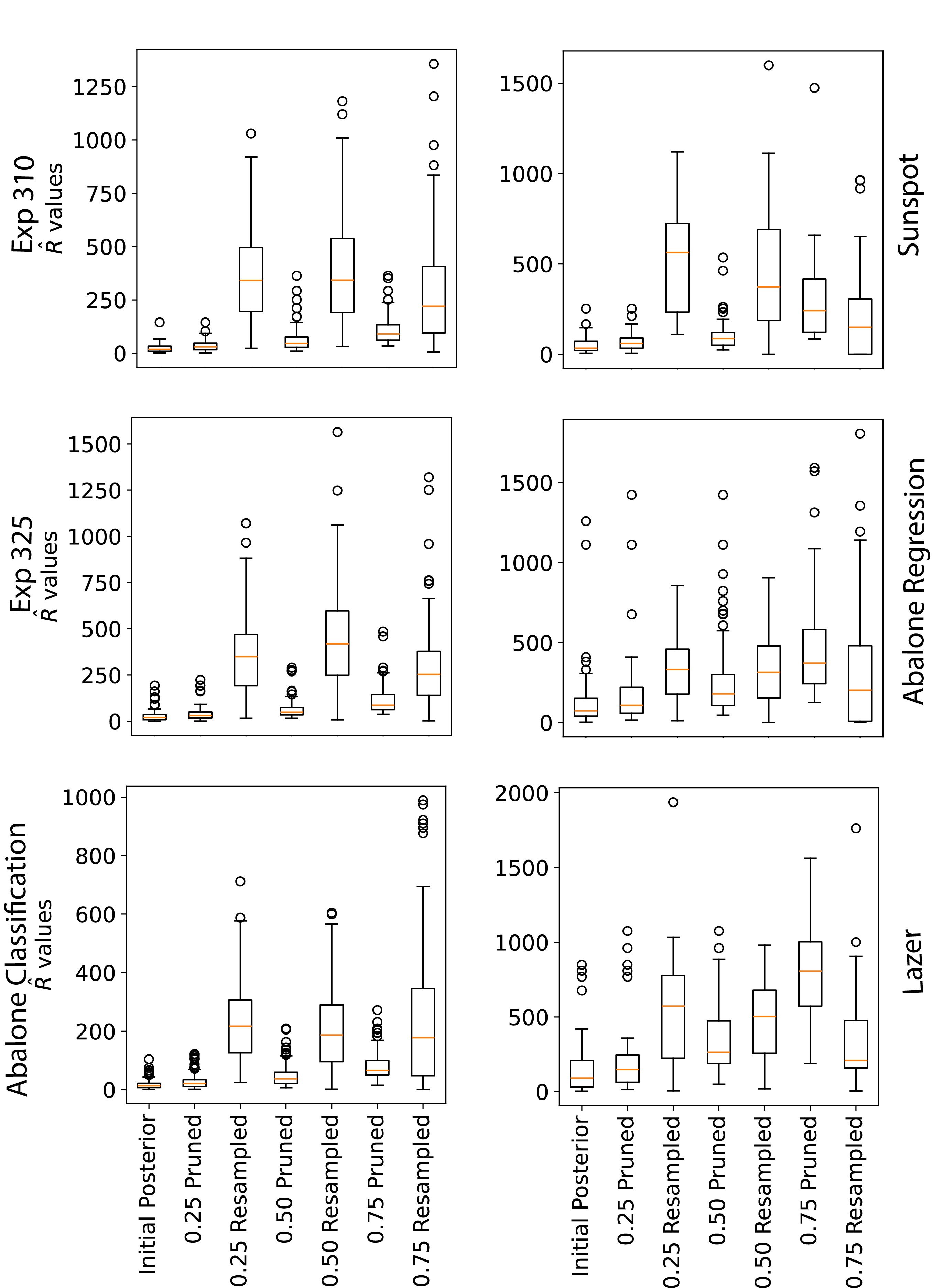}
    \caption{Gelman-Rubin diagnostic values (\(\hat{R}\)) for assessing MCMC sampling convergence across different the selected datasets. We computed the diagnostic score for the initial posterior sampling, and after pruning/resampling at different levels (0.25, 0.50, and 0.75). }
    \label{fig:convergence_test}
\end{figure}

Figure \ref{fig:convergence_test} shows the spread German-Rubin diagnostic values (\(\hat{R}\)) across our testing datasets.   The results indicate that the diagnostic values approach 1.0 across all datasets. A value close to 1 suggests that the chains have mixed well and converged to the target distribution, while values significantly greater than 1 would indicate a lack of convergence. The consistent convergence across datasets demonstrates that the MCMC simulations were appropriately configured, with sufficient burn-in periods, iterations, and suitable prior specifications. We see that post-pruning resampling shows slightly poor convergence with higher (\(\hat{R}\)) values at high pruning rates.  This is an inherent problem with German-Rubin analysis where having fewer samples in the posterior may not show convergence \cite{gelman_bayesian_1995}.  We only resample for 1000 iterations; therefore, the chains do not show convergence as clearly as complete posterior before resampling. However, our trace plot analysis in Figure \ref{fig:traceplot} clearly shows that the models have converged.  

%

\subsection{Reef-Core Application}

\begin{figure}[htb!]
    \centering
    \includegraphics[width=1\linewidth]{ 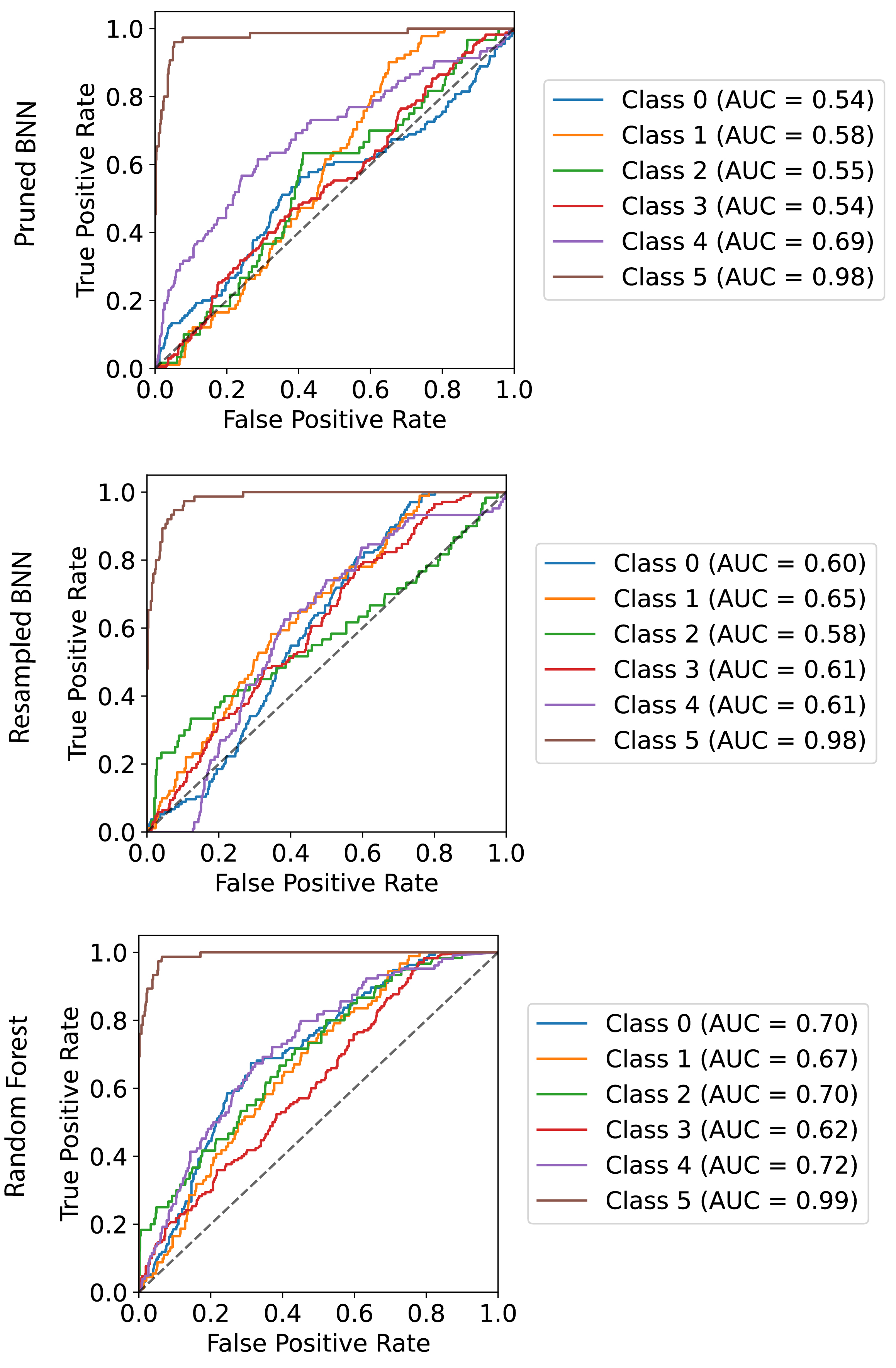}
    \caption{ROC curves for pruned BNN, resampled BNN, and across the six classes of the Expedition 310 dataset. We included results for a Random Forest model for comparison.  Each subplot displays the True Positive Rate (TPR) against the False Positive Rate (FPR) for each class, with corresponding AUC values in the legends. }
    \label{fig:roc_exp310}
\end{figure}

\begin{figure}[htb!]
    \centering
    \includegraphics[width=1\linewidth]{ 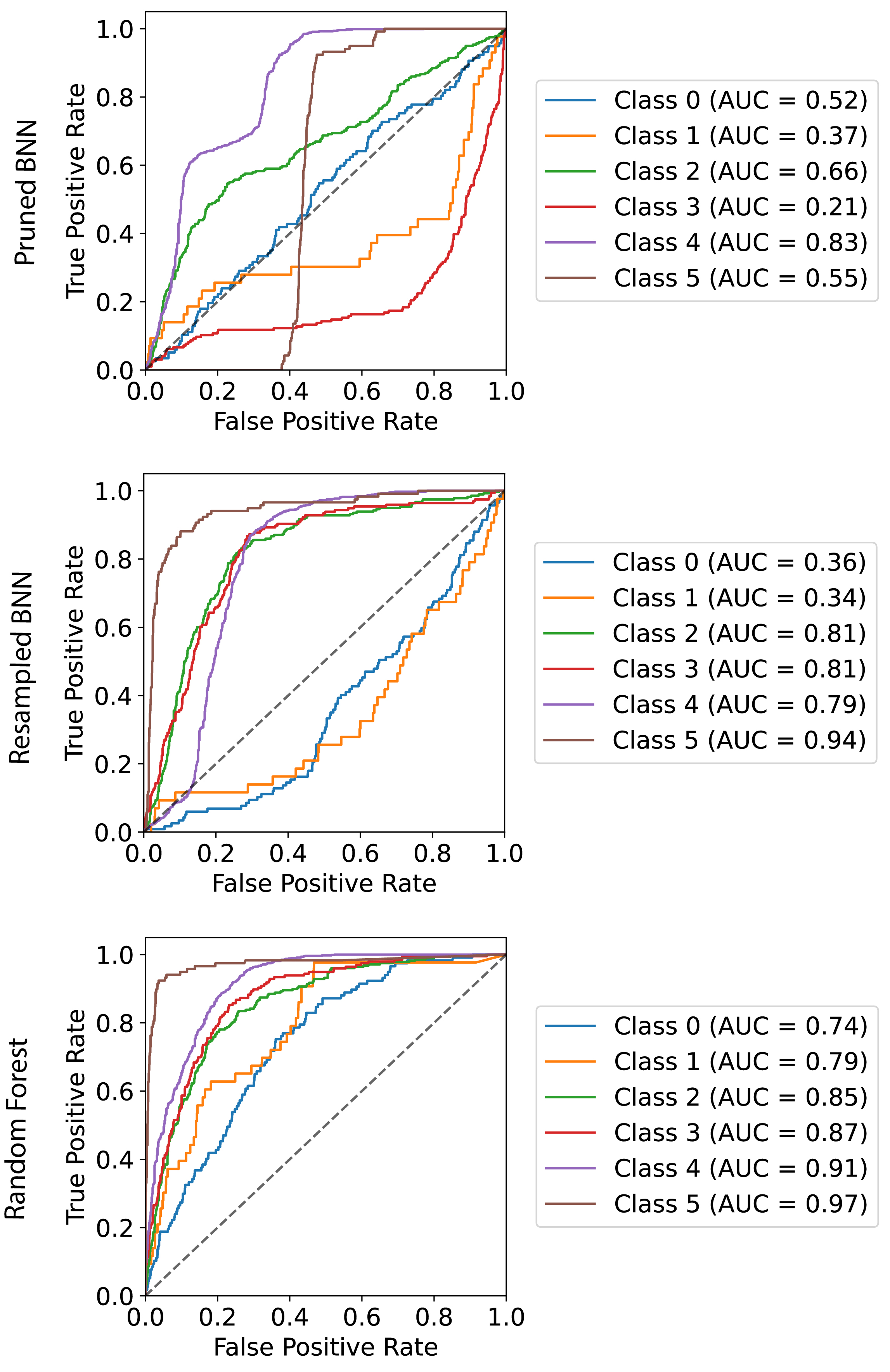}
    \caption{ROC curve and AUC for the six classes in the Expedition 325 dataset, comparing the performance of the pruned BNN, resampled BNN, and Random Forest model.}
    \label{fig:roc_exp325}
\end{figure}

 Figure \ref{fig:roc_exp310} compares the performance of the pruned BNN, pruned and resampled BNN on the multiclass classification task using the Expedition 310 dataset. We include the results from a Random Forest model for comparison, which used 50 model trees.  The pruned BNN exhibited inconsistent performance, with AUC values ranging from 0.54 (Class 0 and Class 3) to 0.98 (Class 5). Although it performs well for Class 5, its predictive capability is limited for other classes, with most AUC scores falling below 0.60. The pruned resampled BNN incorporate resampling post-pruning to improve performance, resulting in moderate improvements for Classes 0, 1, 3, and 4 compared to the pruned BNN. However, its overall performance remains inferior to the Random Forest model, except for Class 5, where its AUC (0.98) is comparable.  Figure \ref{fig:roc_exp325} presents the ROC curve and AUC values  for the three models for the Expedition 325 dataset.  The pruned BNN shows highly variable performance, with AUC values ranging from 0.21 for Class 3 to 0.83 for Class 4. This variability highlights the inconsistency of the pruned BNN in effectively distinguishing between certain classes. Despite this, it achieves reasonable accuracy for some classes (e.g., Class 4). The resampled BNN demonstrates significant improvement over the pruned model, particularly for Classes 2, 3, and 5, where AUC values reach 0.81, 0.81, and 0.94, respectively. In this dataset, we also see that the overall performance remains inferior yet comparable to that of the Random Forest model. We compare our pruned model performance to the Random forest \cite{ho_random_1995} model, as it was the best-performing model for classification in previous work \cite{insua_advanced_2015, deo_reefcoreseg_2024}. The Random Forest model demonstrates superior performance across all six classes, with higher AUC; however, it lacks any uncertainty information about model parameters.  We can see that the BNN methods can keep up with the performance of Random Forests, whilst also providing probabilistic prediction models.

\section{Discussion}
\label{sec:discussion}

The results highlight the significant impact of pruning methods, pruning levels, and resampling on the classification and regression performance of BNNs. Our observations are consistent with prior studies emphasising the superiority of structured pruning methods in maintaining model performance under resource-constrained conditions.

We observed that the structured pruning methods, namely signal-to-noise and signal-plus-noise, consistently outperform random pruning across all datasets and tasks (Table \ref{tab:classification} and \ref{tab:regression}). This advantage is particularly pronounced at higher pruning levels (e.g., 0.75), where random pruning exhibits a substantial drop in classification accuracy and an increase in RMSE (Figure \ref{fig:regression_resample} and \ref{fig:classification_resample}). The effectiveness of structured methods can be attributed to their ability to prioritise the removal of less relevant parameters, thereby preserving critical information within the network. We found that the signal-to-noise pruning method reported the best performance in classification, and signal-plus-noise reported the best performance on regression tasks, indicating their robustness in maintaining model predictive power.

Resampling is a critical factor in mitigating the performance degradation caused by pruning. Across all methods, resampled networks consistently outperform their original counterparts, particularly for random pruning. In classification tasks, resampling improves accuracy significantly for random pruning, narrowing the performance gap with signal-to-noise and signal-plus-noise. Similarly, in regression tasks, resampling improved the BNN prediction accuracy (RMSE) and stabilised performance, even at higher pruning levels.  Models having 75\% of their parameters removed saw upto 25\% gain in performance through resampling.  This demonstrates that resampling effectively leverages the remaining parameters post-pruning, enabling the model to recalibrate and regain its predictive capabilities.

The complexity of the datasets influences the effectiveness of pruning methods. More straightforward datasets, such as Iris and Abalone, show minimal performance degradation across all methods and pruning levels, suggesting that even random pruning retains sufficient information for accurate predictions. In contrast, complex datasets like Exp 310, and Exp 325  are more sensitive to unstructured pruning. In these datasets, we observed that random pruning exhibits significant variability and poorer accuracy, whereas structured methods maintain superior performance. These observations highlight the importance of selecting appropriate pruning strategies tailored to the dataset's characteristics.



 The reef exploration datasets, Exp 325 and Exp 310, presented unique challenges due to their inherent complexity and class imbalance, reflecting marine environments' intricate structure and biodiversity. The structured pruning methods employed in this study, coupled with resampling, showed promise in addressing these challenges by maintaining classification performance and reducing error even at higher pruning levels.  Precise classification whilst modelling uncertainties in drill core analysis is crucial for understanding paleoclimate and assessing climate change impacts \cite{iryu_sea_2010}.  The ability to optimise machine learning models for such resource-constrained platforms could also be applied to datasets such as underwater robotic vehicles and remote sensing systems, offering a pathway for improving the scalability and efficiency of data-driven marine exploration. These analyses often involved labour-intensive tasks with sparse and expensive data, such as those obtained from drill cores \cite{sanborn_new_2017, westphal_genesis_2010} that present significant challenges in the scientific community.  The integration of uncertainty quantification is vital in these applications, allowing researchers to account for the inherent variability in sparse datasets and limited observations. This is particularly relevant when constructing spatio-temporal topography models with uncertainties, such as reef growth models in the Great Barrier Reef \cite{pall_bayesreef_2020}. However, there are limitations to these approaches, including the reliance on priors and the assumptions underpinning them. Improper or overly simplistic priors may result in biased outcomes, especially when dealing with multimodal distributions in small ecological datasets \cite{zhang_effect_2018}.  Addressing these challenges is essential for generating accurate, uncertainty-aware models in drill core and paleoclimate research. These models could also significantly impact marine research, particularly in biota classification, habitat mapping, and long-term environmental monitoring. By integrating structured pruning techniques and resampling, researchers can deploy lightweight yet accurate models for real-time analysis, enabling more effective decision-making in dynamic and resource-limited underwater environments.

 Furthermore, achieving reliable results requires robust convergence analysis, as failure to converge or improper diagnostic evaluation can lead to misleading inferences. Our analysis of the Gelman-Rubin rates for the pruned models has confirmed that the models we have generated have stability in their posterior samples.  Therefore, we can have confidence in our model's predictive performance.  These findings have important implications for optimising deep learning models in resource-constrained environments. The results demonstrate that structured pruning methods, particularly signal-to-noise pruning, should be preferred for both classification and regression tasks, especially when high pruning levels are required (i.e. more compact BNNs). Furthermore, resampling is an essential step to enhance the performance and stability of pruned networks, particularly for unstructured methods like random pruning.


Future research could focus on developing adaptive and dynamic strategies to adjust the pruning thresholds of BNNs based on task requirements and dataset complexity. Additionally, integrating resampling as a standard post-pruning step and exploring iterative pruning-resampling cycles could further enhance the effectiveness and reliability of pruning workflows. Expanding this evaluation to more diverse and complex tasks and datasets could also provide deeper insights into the generalizability of these methods.  The reduced complexity and better convergence of compact BNNs present opportunities for applications in image classification and segmentation tasks, offering robust uncertainty quantification in computer vision. Pruning and resampling strategies can be extended to more complex neural network architectures, including CNNs and deep transfer-learning methods \cite{chandra_bayesian_2021, chandra_bayesian_2020}. There is also potential to bridge the gap between pruning and knowledge distillation \cite{gou_knowledge_2021, yim_gift_2017} by leveraging this Bayesian pruning technique for efficient knowledge transfer in deep network architectures, opening avenues for improved generalisation and model efficiency. Such advancements would facilitate the deployment of computationally efficient, high-performing, and robust deep-learning models in a broad range of real-world applications, including critical domains such as marine ecosystem monitoring and conservation.

\section{Conclusion}
\label{sec:conclusion}

In this study, we systematically investigated the impact of various pruning methods, pruning levels, and the inclusion of resampling on the performance of BNNs across multiple classification and regression tasks. The results demonstrated that structured pruning methods, namely signal-to-noise and signal-plus-noise, significantly outperformed random pruning, particularly at higher pruning levels. The superior performance of structured methods is attributed to their ability to prioritise the removal of less relevant parameters while preserving those critical for accurate predictions. In contrast, random pruning resulted in substantial performance degradation, particularly in more complex datasets, highlighting its inefficiency as a pruning strategy.

The incorporation of resampling after pruning substantially enhanced the performance accuracy of pruned networks by enabling recalibration of the remaining parameters. Resampling proved particularly effective in mitigating the adverse effects of unstructured pruning, consistently improving classification accuracy and reducing regression error across all datasets. This underscores the necessity of integrating resampling into pruning workflows, especially for methods prone to performance variability.  Robust convergence diagnostics, through Gelman-Rubin analysis, confirmed better convergence of pruned BNNs for regression tasks at high pruning levels, further highlighting the importance of structured pruning methods. 

The reef exploration datasets, Exp 325 and Exp 310 highlighted challenges such as class imbalance and data sparsity inherent in marine environments. Pruning and resampling techniques effectively maintained classification accuracy, while modelling uncertainties in drill core analysis, essential for paleoclimate research and climate impact assessment. Future work should focus on dynamic pruning strategies and iterative pruning-resampling cycles, extending these methods to complex neural architectures like CNNs and deep transfer-learning models. This could facilitate the use of probabilistic modelling in real-world applications, improving scalability, robustness, and generalisation whilst providing uncertainty in predictions. 

\section{Data and Code Availability}
The code and data used for implementing the framework for Bayesian Pruning are available on the GitHub repository 
\url{https://github.com/rvdeo/bayes_prun}

\subsubsection {Declaration of generative AI and AI-assisted technologies in the writing process}

During the preparation of this work, the authors used GPT 4.0 in order to assist in evaluating the grammatical correctness of the text. After using this tool/service, the authors reviewed and edited the content as needed, and take full responsibility for the content of the publication.

\bibliographystyle{elsarticle-num}
\bibliography{main} 

\section{Appendix}

\begin{figure*}[htb!]
    \includegraphics[width=1\linewidth]{ 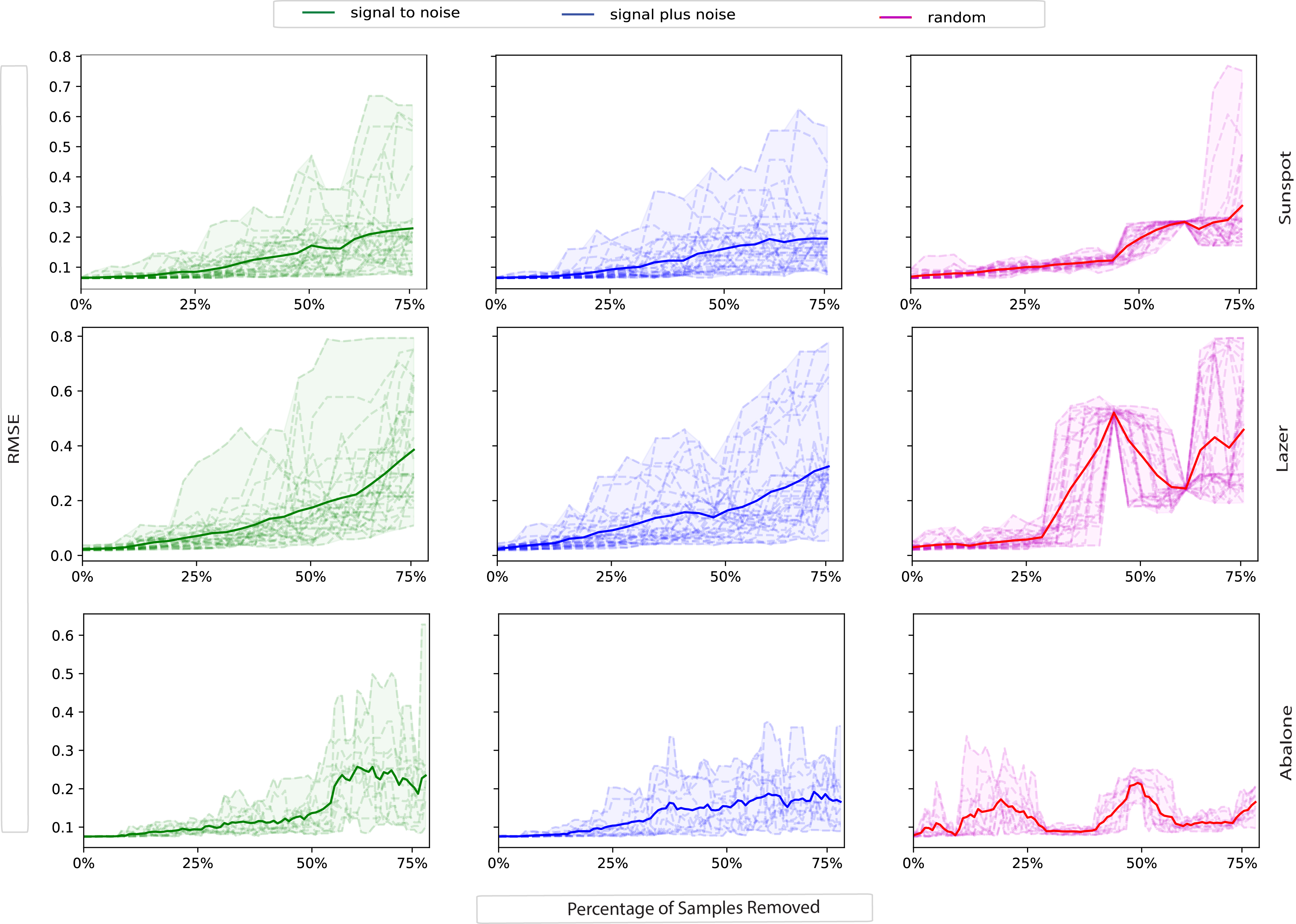}
    \caption{Post pruning regression performance over 30 experimental runs on all regression datasets.  We show the changes to each model's posterior prediction performance with 0 to 75 \% pruning using each pruning scheme. The bold lines in each plot represent the mean RMSE across the experimental runs.}
    \label{fig:regression_error_variability}
\end{figure*}

\begin{figure*}[htb!]
    \centering
    \includegraphics[width=\textwidth]{ 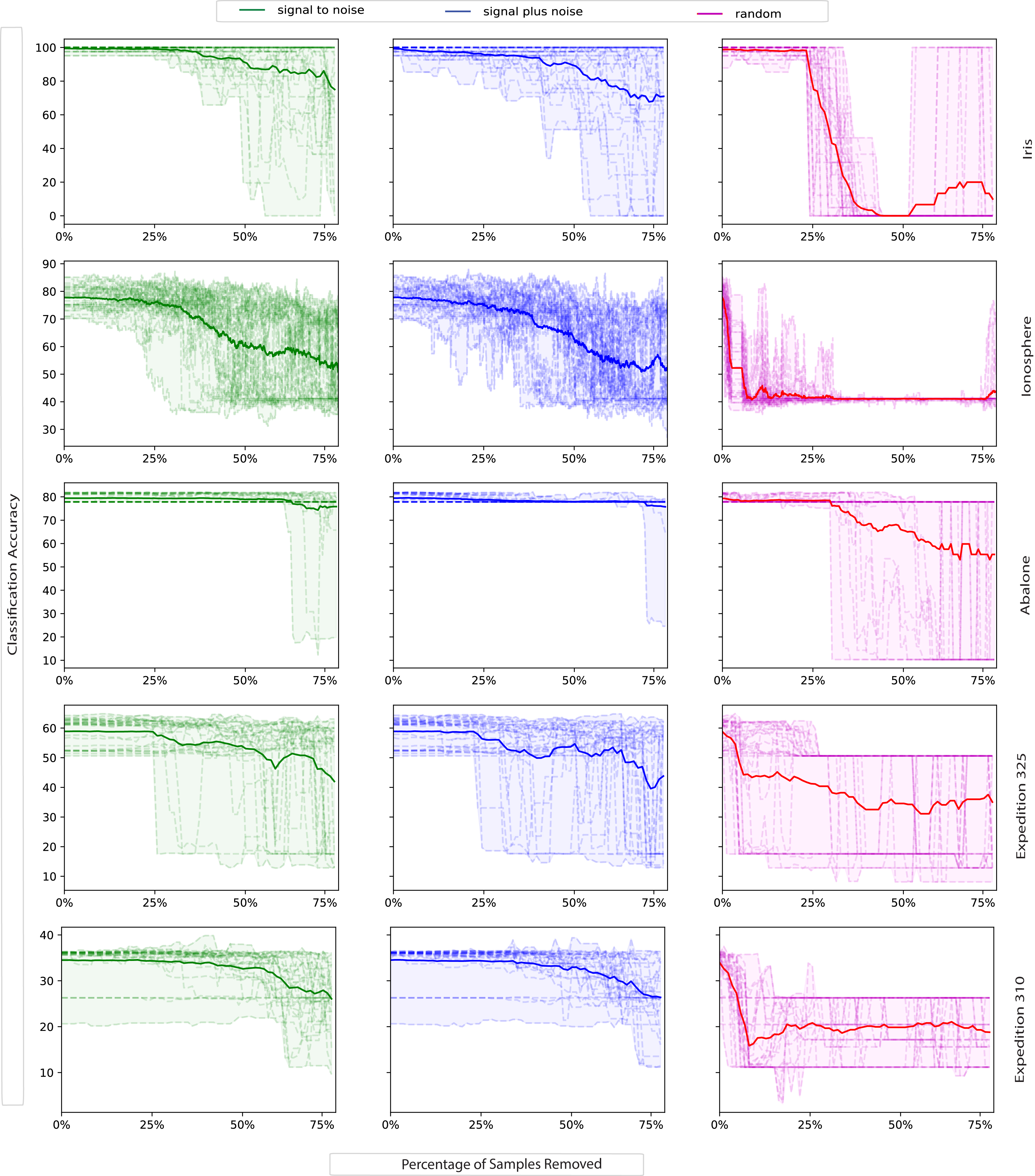}
    \caption{Post pruning classification performance over 30 experimental runs on all classification datasets.  We show the changes to each model's posterior prediction performance with 0 to 75 \% pruning using each pruning scheme. The bold lines in each plot represent the mean classification accuracy across the experimental runs.}
    \label{fig:classification_error_variability}
\end{figure*}



\end{document}